\documentclass[lettersize,journal]{IEEEtran}
\usepackage[caption=false,font=normalsize,labelfont=sf,textfont=sf]{subfig}
\usepackage{textcomp}
\usepackage{stfloats}
\usepackage{url}
\usepackage{verbatim}
\usepackage{graphicx}
\usepackage{cite}
\usepackage{graphicx}
\usepackage{amsmath}
\usepackage{amssymb}
\usepackage{booktabs}
\usepackage[ruled,vlined]{algorithm2e}
\usepackage{algpseudocode}
\usepackage{pdflscape}
\usepackage{booktabs}
\usepackage{multirow}
\usepackage{amsmath}
\usepackage{mathtools,xparse}
\usepackage{hyperref}
\usepackage[table]{xcolor}

\definecolor{lightgray}{gray}{0.9} % Adjust the shade by changing the number (0.9 in this case)
\definecolor{lightblue}{RGB}{173, 216, 230}

\definecolor{blue1}{RGB}{173, 216, 230}  % Very light blue, use dark text
\definecolor{blue2}{RGB}{135, 206, 235}  % Sky blue, use dark text
\definecolor{blue3}{RGB}{100, 181, 205}  % Clear blue sky, use white text
\definecolor{blue4}{RGB}{70, 130, 180}   % True medium blue, use white text
\definecolor{blue5}{RGB}{0, 102, 204}    % Deep blue, use white text
\definecolor{blue6}{RGB}{0, 51, 202}     % Navy blue, use white or light gray text
\definecolor{blue7}{RGB}{0, 51, 102}     % Navy blue, use white or light gray text

\begin{document}

\title{Dataset Condensation Driven Machine Unlearning}

% \author{Junaid Iqbal Khan\thanks{Email: \texttt{email@example.com}}}
\author{\IEEEauthorblockN{Junaid Iqbal Khan}
\IEEEauthorblockA{\\
Email: dianujkotov15@gmail.com}}
        % <-this % stops a space



\maketitle

\begin{abstract}
The current trend in data regulation requirements and privacy-preserving machine learning has emphasized the importance of machine unlearning. The naive approach to unlearning training data by retraining over the complement of the forget samples is susceptible to computational challenges. These challenges have been effectively addressed through a collection of techniques falling under the umbrella of machine unlearning. However, there still exists a lack of sufficiency in handling persistent computational challenges in harmony with the utility and privacy of unlearned model. We attribute this to the lack of work on improving the computational complexity of approximate unlearning from the perspective of the training dataset.
In this paper, we aim to fill this gap by introducing dataset condensation as an essential component of machine unlearning in the context of image classification. To achieve this goal, we propose new dataset condensation techniques and an innovative unlearning scheme that strikes a balance between machine unlearning privacy, utility, and efficiency. Furthermore, we present a novel and effective approach to instrumenting machine unlearning and propose its application in defending against membership inference and model inversion attacks.
Additionally, we explore a new application of our approach, which involves removing data from `condensed model', which can be employed to quickly train any arbitrary model without being influenced by unlearning samples. The corresponding code is available at \href{https://github.com/algebraicdianuj/DC_U}{URL}.
\end{abstract}


%%%%%%%%%%%%%%%%%%%%%%%%%%%%%%%%%%%%%%%%%%%%%%%%%%%%%%%%%%%%%%%%%%%%%%%%%%%%%%%%%%%%%%%%%%%%%%%%%
\begin{figure*}[t]
\centering
\hspace*{-0.1in}
\includegraphics[width=18.5cm, height=5.8cm]{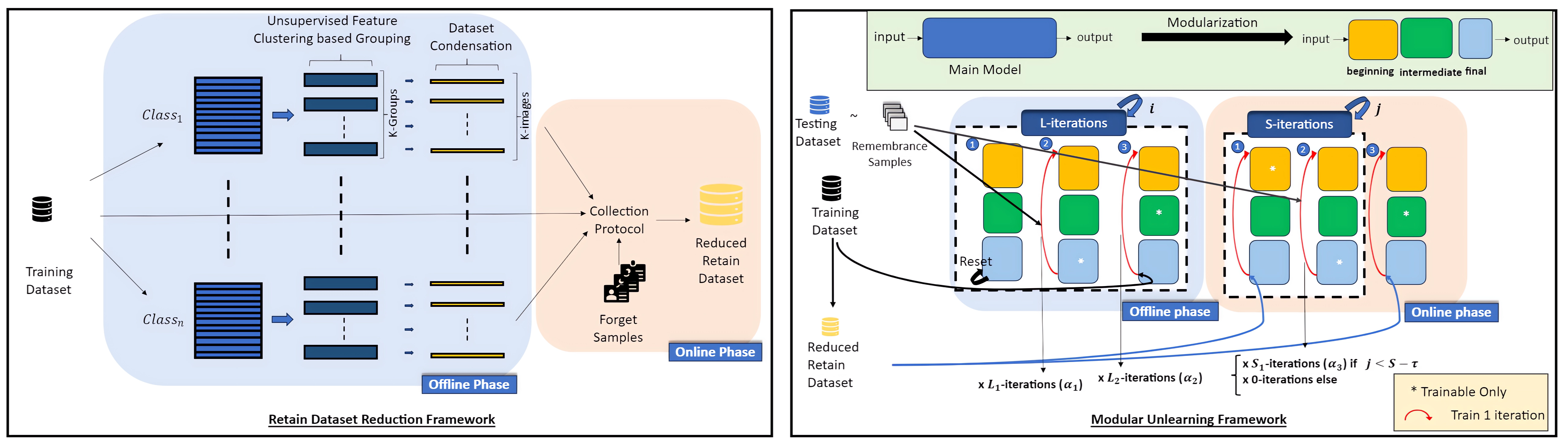}
\caption{Main abstraction of Proposed Scheme}
\end{figure*}
%%%%%%%%%%%%%%%%%%%%%%%%%%%%%%%%%%%%%%%%%%%%%%%%%%%%%%%%%%%%%%%%%%%%%%%%%%%%%%%%%%%%%%%%%%%%%%%

\section{Introduction}

% \IEEEPARstart{T}{he} 
The significance of the machine unlearning has already been established and well described in relationship to international data regulations like the ‘the right to be forgotten’ \cite{rosen2011right} clause in General Data Protection Regulation (GDPR) \cite{hoofnagle2019european}. Besides the main task of removing the user data from model unlearning, machine unlearning has found applications in other areas of privacy preserving machine learning like mitigating bias \cite{kurmanji2023towards},  mitigating backdoor attacks \cite{liu2022backdoor} etc. On the other hand, while this topic is still in its infancy, machine unlearning as privacy solution as been studied to be vulnerable to other kind of privacy attacks \cite{marchant2022hard,di2022hidden}. In any case, any unlearning algorithm is supposed to design to compete with the effects of naive unlearning approach of retraining on the remaining dataset (not including the samples to be forgotten) from scratch, but with additional caveat that the designed unlearning algorithm should be much more efficient. This target has been extensively studied to be achieved under classifications of `exact machine unlearning' and `approximate machine unlearning', where the associated technique exactly or approximately mimic the effects of naive unlearning, respectively. Its sister approach, namely `catastrophic forgetting' \cite{feng2022overcoming}, which involves fine-tuning a pre-trained model over subset of the training dataset, and the model starts under performing upon the compliment of that subset. However, catastrophic forgetting has been treated more a challenges in machine learning, especially in incremental learning, than a commodity. However, the techniques under `approximate machine unlearning' have been gaining much more popularity since they are computationally much more efficient than their exact unlearning counterparts and have been shown to be successful in approaching the metrics of naive unlearning.  
\\
Despite the popularity of approximate machine unlearning algorithms, they suffer from high margin between efficiency (the amount of time for unlearning algorithm completion), privacy (protection against adversary to infer the forgotten data from unlearned model) and utility (preservation of performance of unlearned model on retain dataset). One important illustration of this challenge is the work done in \cite{koh2017understanding}, where a close form expression for difference between original and unlearning parameters is derived, assuming the distance between them is sufficiently small, but leads to a computationally expensive solution involving Hessians, which may not be applicable at all for large models. 
If we take step back, and focus on the potential of other domains of machine unlearning within approximate machine unlearning, then there have been several techniques studied to be beneficial for utility and efficacy perspective of unlearning, like distillation \cite{chundawat2023can, kurmanji2023towards}, model pruning \cite{jia2023model} etc. Until now, this line of work has been centered around the model perspective. In other words, the unlearning algorithm have predominantly focused on modifying either the model's loss function or its parameters. In this paper, we take a digression and focus on dataset, as well as, model centeric machine unlearning scheme, which aims to fill the gap in unlearning literature to find a good median between privacy, utility and efficiency of approximate unlearning algorithm. More specifically, we design new dataset condensation techniques to reduce the training fodder for unlearning, and new unlearning scheme, which we term `modular unlearning' to further accelerate unlearning via catastrophic forgetting. To simply describe the modularized training, we essentially split the model into three parts and train them seperately, the consequence of which is that middle part requires minimum epochs to achieve catastrophic forgetting. We also metricize this unlearning in two new ways, namely via `unlearning' and `overfitting' metric. Lastly, we envision our algorithm towards two new and important applications.
\\We summarize our major contributions as follows:
\begin{itemize}
    \item We propose two new dataset condensation techniques as means to reduce the size of compliment of forget samples (retain dataset) for the training part of unlearning.
    \item We propose modularized unlearning, focused toward image classification task, which splits the pre-trained model into three parts and seperately trains them using the reduced retain dataset.
    \item We propose two new metrics to measure unlearning, namely `unlearning' and `overfitting' metric.
    \item We propose two new applications of proposed unlearning. First one provides the defense of membership inference attack as a competitor of differentially private (DP)-Adam based training \cite{bu2022differentially1,bu2022differentially2}. Second one allows removal of information from forget samples from an autoencoder, which when augmented with any new model can lead to fast learning.
    \item We conduct extensive experiments and show that our unlearning methodology finds a good balance between unlearning privacy, utility and efficiency, as compared to state of the art approximate unlearning approaches.
\end{itemize}

\section{Related Works}
% \subsection{Machine Unlearning}
The problem of \textbf{machine unlearning}, i.e. to find a fast alternative to naive retraining, is non-trivial. For example, an obvious approach of gradient ascent over forget samples can quickly fail \cite{thudi2022unrolling}. The current machine unlearning algorithms can be broadly divided into `exact' and `approximate' machine unlearning algorithms.
\\
% It might be intuitive to just perform gradient ascent on the loss of forget samples, but the moving towards local maxima of loss over forget samples, does not imply that same solutions in parameter space are the local minima of loss over retain samples, the evidence of which can be found in \cite{thudi2022unrolling}. In realization of fault in this obvious route, 
Exact machine unlearning attempts to emulate retraining, but in an optimized manner. However, the collection of techniques are still pertinent to computational and scalability challenges. One important work in this regard is partitioning of datasets into multiple subsets, which are themselves partioned as well. This is followed by the individual training of independent models on each discrete subset, and the outputs are ensembled \cite{bourtoule2021machine}. The first machine unlearning work \cite{cao2015towards} also falls with in this abstraction, where by converting the machine learning system to summation form, the unlearning request updates few of the summation terms.
\\
Approximate unlearning algorithm achieve either a certified or a heuristically justifiable approach to achieve the effects of naive retraining with significant efficiency advantage. One of the main strategies in this regard is to be parameter focussed, and to either subtract the parameter updates due to batch per epoch gradient updates from forget samples in the prior training scheme \cite{graves2021amnesiac, wu2020deltagrad, thudi2022unrolling} or performing single step updates via gradients \cite{warnecke2021machine} or Hessians \cite{golatkar2020eternal, mahadevan2021certifiable}. Another important and rather ubiquitous strategy is to focus on training (fine-tuning) on forget dataset to achieve good unlearning privacy evaluations, and train on retain dataset to achieve competitive unlearning utility metrics \cite{tarun2023fast,kurmanji2023towards,chundawat2023can}. A recent trend has been to find intersection of other branches of deep learning like adversarial attacks \cite{chen2023boundary}, model sparsity/ pruning \cite{jia2023model} and model distillation \cite{kurmanji2023towards,chundawat2023can}, with unlearning, which are shown to be promising as in improving the per-epoch unlearning capacity as compared to naive retraining, and thus in this way, also improving upon unlearning efficiency. Model privacy is mostly metricized via membership inference attack \cite{shokri2017membership,kurmanji2023towards,jia2023model}, which is the simplest attack in privacy-preserving machine literature that aims to infer the probability whether a particular sample was using in training of model or not. Another way that has been shown to depict the unlearning privacy is via model inversion attack \cite{fredrikson2015model, chundawat2023zero}, where one attempts to reconstruct the training data using the trained model.

% \subsection{Dataset Condensation}
The research question of whether large dataset can be reduced into smaller samples (so as to say condensing dataset), which then trained on arbitrary model would lead to similar accuracy as that of the original dataset, has been of great interest in recent years.
% In this regard, while there exist a family of techniques like coreset selection methods \cite{guo2022deepcore,aljundi2019gradient} with similar goal, \textbf{dataset condensation} (or dataset distillation) is shown to maximize the information of whole dataset to small number of synthetic samples in a model driven manner.
Under the umberella of dataset condensation, the techniques for condensing dataset solely rely on convex optimization of the random images, such that either gradient of model trained on them and on the original dataset \cite{zhao2020dataset}, or the distance between distribution of pretrained model's features on them and the original dataset \cite{zhao2023dataset}, or the distance between parameter states over training trajectories when trained on them and on original dataset \cite{cazenavette2022dataset}, is minimized. Whilst there have been several improvements upon these strategies \cite{zhao2021dataset, zhao2023improved, kim2022dataset}, a persistent major hurdle in their quick rapid adoption for downstream deep learning applications, including unlearning, is the associated computational bottleneck.

\begin{algorithm}
\footnotesize
\caption{Unsupervised feature clustering (K-means) based grouping}
\KwIn{Training images $T= \bigcup_{i=1}^{c}{\bigcup_{j=1}^{n}}T_{ij}$, and a pre-trained network $\mathcal{M}$}
\KwOut{Image clusters $\bigcup_{i=1}^{c}{\bigcup_{j=1}^{K}}\Gamma_{ij}$}
Image Clusters $C=\{\}$\;
\ForEach{class $i$}{
    $F_i=\mathcal{M}_\text{features}(\bigcup_{j=1}^{n}T_{ij})=\bigcup_{j=1}^{n}\mathcal{M}_\text{features}(T_{ij})$\;

    Perform K-means clustering on $F_i$ resulting in clusters ${\bigcup_{j=1}^{K}}\Gamma_{ij}$ of the training images\;

    $C=C\cup{\bigcup_{j=1}^{K}}\Gamma_{ij}$\;
}
\Return{C}
\end{algorithm}

\begin{algorithm}
\footnotesize
\caption{Dataset Condensation via Fast Distribution Matching}
\KwIn{Image Clusters $C = \bigcup_{i=1}^{c}{\bigcup_{j=1}^{K}}\Gamma_{ij}$, epochs $E$}
\KwOut{Condensed Images $\mathcal{C}=\bigcup_{i=1}^{c}{\bigcup_{j=1}^{K}}\phi_{ij}$}
$\mathcal{C}=\{\}$\;
\ForEach{cluster $\Gamma_{ij}$}{
    Initialize weighted average function $W$ with parameters $\omega\in\mathbb{R}^{|\Gamma_{ij}|}$\;
    $\text{epoch}=1$\;
    \While{$\text{epoch}\leq{E}$}{
        $\mathcal{L}=\|\text{mean}(\mathcal{M}_\text{features}(\Gamma_{ij}))-\text{mean}(\mathcal{M}_\text{features}(W(\Gamma_{ij})))\|_2$\;
        Optimize $\omega$ by backpropagating the $\mathcal{L}$\;
    }
    $\phi_{ij}=W(\Gamma_{ij})$\;
    $\mathcal{C}=\mathcal{C}\cup{\phi_{ij}}$\;
    $\text{epoch}=\text{epoch}+1$\;
    
}
\Return{$\mathcal{C}$}
\end{algorithm}

\section{Preliminaries and Notation}

We define the original dataset as $\mathcal{D}=\{T_i, l_i\}_{i=1}^{N_D}$,  where $T_i$ is the $i^{th}$ training image and $l_i$ is the $i^{th}$ label. $\mathcal{D}$ can be partitioned into the forget dataset $\mathcal{F}=\{T_{f_i}, l_{f_i}\}_{i=1}^{N_F}$ and retain dataset $\mathcal{R}=\{T_{r_i}, l_{r_i}\}_{i=1}^{N_R}$. Here, $\mathcal{F} \cup \mathcal{R} = \mathcal{D}$, and $\mathcal{F} \cap \mathcal{R} = \emptyset$. For the sake of representation, we represent the images in $\mathcal{D}$ as $T = \bigcup_{i=1}^{c} \bigcup_{j=1}^{n} T_{ij}$, where $c$ is the total number of classes in $\mathcal{D}$, $n$ is the number of images per class such that $N_D=nc$ and $T_{ij}$ represents the image in $i^{th}$ class and $j^{th}$ index.
\\
% The forget set $\mathcal{F}$ represents the part of the training dataset $\mathcal{D}$ to be forgotten by a model, trained on $\mathcal{D}$ using loss function $\mathcal{L_{\text{CE}}}$. Lets call this trained model as $\mathcal{M}_{\theta}$ with trained parameters $\theta$. The goal of dataset reduction framework is reduce $\mathcal{R}\rightarrow{\mathcal{R}_{\text{red}}}$, such that $|\mathcal{R}_{\text{red}}|<|\mathcal{R}|$. Within the dataset reduction framework, the images of each class are grouped into $K$ clusters, such that $K<n$. Finally, after unlearning procedure, the model \mathcal{M} gains parameters $\theta^{*}$.
The forget set $\mathcal{F}$ represents the part of the training dataset $\mathcal{D}$ to be forgotten by a model, trained on $\mathcal{D}$ using loss function $\mathcal{L_{\text{CE}}}$. Let's call this trained model as $\mathcal{M}_{\theta}$ with trained parameters $\theta$. The goal of the dataset reduction framework is to reduce $\mathcal{R}\rightarrow{\mathcal{R}_{\text{red}}}$, such that $|\mathcal{R}_{\text{red}}|<|\mathcal{R}|$. Within the dataset reduction framework, the images of each class are grouped into $K$ clusters, such that $K<n$. Finally, after the unlearning procedure, the model $\mathcal{M}$ gains parameters $\theta^{*}$.

\section{Methodology}
In our methodology, we propose two frameworks. The first framework provides the minimum amount of training data for unlearning, in form of reduced retain dataset. The second framework performs the unlearning by using reduced retain dataset. Both of these frameworks have an offline and online phase, where the offline phase happens prior to actual unlearning phase, in a reasonably fast manner. The online phase happens during each unlearning cycle.

\subsection{Retain Dataset Reduction Framework}
This framework comprises of an offline and online phase. The offline phase condeses the whole training dataset $D$ into condensed form. During online phase, which happens during each unlearning cycle, the collection protocol takes in the condensed training dataset, and forget dataset $\mathcal{F}$ to filter out a reduced dataset $\mathcal{R}_{\text{red}}$.

\begin{algorithm}
\footnotesize
\caption{Image Condensation via Model Inversion}
\KwIn{Image Clusters $C = \bigcup_{i=1}^{c}{\bigcup_{j=1}^{K}}\Gamma_{ij}$ with individual cluster labels $l_{ij}$ for  cluster images $I_{ij}$ with original labels $l_{i}$, epochs $E$, regularization parameter $\lambda$ and pre-trained network $\mathcal{M}$}
\KwOut{Condensed Images $\mathcal{C}=\bigcup_{i=1}^{c}{\bigcup_{j=1}^{K}}\phi_{ij}$}
Create InverterNet $\Lambda: l_{ij}\rightarrow{\Gamma_{ij}}$ with parameters $\theta_{\Lambda}$\;
Make $\mathcal{M}$ parameters $\theta_{\mathcal{M}}$ untrainable\;
Compose $\Lambda$ and $\mathcal{M}$ as $\mathcal{M}\Lambda: l_{ij}\rightarrow{l_i}$\;
$\text{epoch}=1$\;
\While{$\text{epoch}\leq{E}$}{
    $\mathcal{L}=\mathcal{L}_{\text{CE}}(\mathcal{M}\Lambda(l_{ij}),l_i)+\lambda\mathcal{L}_{\text{MSE}}(\Lambda(l_{ij}),\Gamma_{ij})$\;
    Optimize $\theta_{\Lambda}$ by backpropagating the $\mathcal{L}$\;
    $\text{epoch}=\text{epoch}+1$\;
}
$\mathcal{C}=\{\}$\;
\ForEach{cluster label $l_{ij}$}{
$\phi_{ij}=\Lambda(l_{ij})$\;
$\mathcal{C}=\mathcal{C}\cup{\phi_{ij}}$\;
}
\Return{$\mathcal{C}$}
\end{algorithm}

\subsubsection{Offline Phase}

For each $i^{th}$ class in the dataset $D$, the images $\bigcup_{j=1}^{n}T_{ij}$ is grouped into $\bigcup_{j=1}^{K}\Gamma_{ij}$ using algorithm-1, where $\Gamma_{ij}$ represents a $j^{th}$ cluster of images for the $i^{th}$ class, and $|\Gamma_{ij}|=\frac{n}{K}$ (assuming that the clustering algorithm leads to clusters of equal sizes). For each cluster $\Gamma_{ij}$, we assign cluster label $l_{ij}=j{l_i}$, such that each image in $\Gamma_{ij}$ has label $l_{ij}$. Over all clusters, i.e. $\bigcup_{i=1}^{c}{\bigcup_{j=1}^{K}}\Gamma_{ij}$, we condense each cluster into a single image, either via our proposed fast distribution matching and model inversion.

\begin{algorithm}
\footnotesize
\caption{Collection Protocol}
\KwIn{Image clusters $C = \bigcup_{i=1}^{c}{\bigcup_{j=1}^{K}}\Gamma_{ij}$, condensed images $\mathcal{C}=\bigcup_{i=1}^{c}{\bigcup_{j=1}^{K}}\phi_{ij}$, forget dataset $\mathcal{F}=\{T_{f_i}, l_{f_i}\}_{i=1}^{N_F}$
}
\KwOut{Reduced retain dataset $\mathcal{R}_{\text{red}}$}
$\mathcal{R}_{\text{red}}=\{\}$\;
\ForEach{forget image $T_{f_i}$}{
\ForEach{cluster $\Gamma_{ij}$}{
        \If{$T_{f_i}\notin \Gamma_{ij}$}{
            $\mathcal{R}_{\text{red}}=\mathcal{R}\cup({\phi_{ij}},l_i)$\;
        }
        \Else{
            $\mathcal{R}_{\text{red}}=\mathcal{R}\cup(\Gamma_{ij}\setminus{T_{f_i}},l_i)$\;
        }
        }
}
\Return{$\mathcal{R}$}
\end{algorithm}

\subsubsection{Dataset Condensation via Fast Distribution Matching}
Contrary to original distribution matching approach based condensation \cite{zhao2023dataset} where images were optimized, we focus on optimizing the weighted average of images in cluster $\Gamma_{ij}$, leading to drastically low trainable parameter count $\frac{n}{K}$ contrary to parameter count as $n$ times the product of dimensions of training images, as in \cite{zhao2023dataset}.
The images within each $\Gamma_{ij}$ is condensed into a single image through a trainable weighted average, and the weights are optimized by matching the mean of distribution of features associated with condensed image, and the original image, assuming that features follow a gaussian distribution. This process is summarized in algorithm-2.

\subsubsection{Dataset Condensation via Model Inversion}
In model inversion attack \cite{fredrikson2015model}, we essentially find the inverse mapping for a given model, which can map output to input. Inspired by this, we create a new deep learning model, called `InverterNet' $\Lambda$, for the given pretrained model $\mathcal{M}$, such that the composition of these models maps $l_{ij}$ to $l_i$, i.e., $\mathcal{M}\Lambda: l_{ij}\rightarrow l_i$. It should be noted that if $K=1$, then $\mathcal{M}\Lambda$ is an identity mapping, and $\Lambda$ would form an inverse mapping of $\mathcal{M}$. Afterwards, the composition $\mathcal{M}\Lambda$ is trained on the cluster labels, original labels, and original images, such that the standard cross-entropy loss is regularized with the reconstruction error of the InverterNet from the original images. After training, the condensed images with respect to each label $l_i$ is collected by getting outputs of $\Lambda$ over $l_{ij}$. This procedure is described in algorithm-3.

\subsection{Online Phase}
In the online phase, the unlearning requests are aggregated to form forget dataset. Through the collection protocol acting on forget dataset, image clusters and the condensed dataset, we retrieve the reduced retain dataset for unlearning scheme of size $N_r$, whose size is much smaller than the original retain dataset, in time complexity equivalent to that of retrieving original retain dataset.

In collection protocol, if the forget dataset images are not found in image cluster $\Gamma_{ij}$, then the corresponding condensed image with label is collected. Otherwise, we collect the residual retain dataset images with in the cluster, other than the forget dataset images. If we assume that the number of images in all $cK$ clusters is same, then under assumption that forget samples are randomly distributed through out the dataset, and thus through clusters, one can develop as asymptotic bound of the compression ratio of reduced retain dataset, defined as $\eta_T=\frac{N_r}{N_R}$, through application of the collection protocol as.
\begin{equation} \label{eq1}
\eta_T=(1-\frac{1}{cK})^{N_D-cK}\frac{cK}{N_R}+\mathcal{O}((\frac{N_D}{N_R}-1)(1-\frac{1}{cK})(\frac{N_D}{cK}-1))
\end{equation}
Under the same assumptions, we can determine that for $\eta_T<1$, then following inequality to hold.
\begin{equation} \label{eq1}
N_R>N_D-cKlog(cK)+1
\end{equation}

These assumptions are only valid if the unsupervised clustering algorithm partitions data into $K$ clusters of equal sizes. For the adoption of this this work, we only restricted to k-means clustering which partitions data into K-clusters of possibly unequal sizes, based on distance in feature domain. On the other hand, there exists techniques like startified K-means clustering functions to partition into clusters of equal sizes.

\subsection{Modular Training}

In neural networks, especially in convolution ones \cite{wiatowski2017mathematical}, the output of layers progressively becomes translation-invariant from shallow to deep layers. This implies that shallower layers have relatively more information of input \cite{dosovitskiy2016inverting, mo2020darknetz}, thus more vulnerable to model inversion attack. On the other hand, the deeper layers \cite{nasr2018comprehensive, mo2020darknetz} have more information of output, and thus more vulnerable to membership inference attack. In light of this observation, we attempt to partition a neural network into three parts, by grouping the layers from shallow to deeper part into compartments, namely \textbf{beginning}, \textbf{intermediate} and \textbf{final} respectively. These compartments are trained separately in a systematic manner, which we call `modular training', to achieve certain privacy and efficiency goals. This partitioning is depicted in figure-1. We devise an offline and online phase to modular training to achieve unlearning.

\subsubsection{Offline Phase}
In the offline phase, we first sample $M$ images per label from the testing set and call them remembrance samples, where $M$ is small, e.g. 1-10. Then in 
the each of $R$ iterations, we follow three steps. In the first step, we reset the parameters of \textbf{final} to original weights as that of pretrained model $\mathcal{M}$. In second step, the \textbf{final} is only kept trainable, and trained over remembrance samples. In third step,  only \textbf{intermediate} is kept trainable and trained on the original dataset, where the \textbf{beginning}, being dense in information of training images, acts a feature extractor to \textbf{intermediate}. 
\\
The remembrance training over \textbf{final} has two main objectives. First we reduce the vulnerability to membership inference attack to deeper layers \cite{nasr2018comprehensive}, as previously mentioned. Secondly, we induce the application of the resultant neural network towards a situation that, if the parameters of \textbf{final} or even its architecture is arbitrarly changed, then training only the \textbf{final} on remembrance samples (which are quite few) can regain back the accuracy.

\subsubsection{Online Phase}
In online phase, which actively performs unlearning, we assign $S$ iterations, in which we perform training of \textbf{beginning} and \textbf{final} in two steps. In the first step, we train the \textbf{beginning} only for 1 epoch on reduced retain dataset, while in second step, we train \textbf{final} on remembrance samples for $S_1$ iterations.  We introduce a condition to perform second step, if the current iteration of $S$ iterations is less than $S-\tau$, where $0<\tau<S$ is a hyperparameter, which is designed to not degrade the accuracy of retain dataset, in later stage of $S$ iterations. By training the \textbf{beginning}, we are generally reducing the FLOPs associated with training, since the gradient of loss of neural network is highly sensitive to gradient of shallow layers. After $S$ iterations, we perform 1 step training of \textbf{intermediate} over reduced retain dataset. Because in the offline phase, the \textbf{intermediate} was trained with \textbf{beginning} acting as the feature extractor, the training knowledge was attempted to  concentrate in \textbf{intermediate}. By re-modifying the feature extractor, i.e. \textbf{beginning}, the knowledge of \textbf{intermediate} is rendered obsolete, and thus with even 1 iteration, there is immediate catastrophic forgetting. This effect has been seen in transfer learning \cite{raghu2019transfusion} frequently, but never attributed to catastrophic forgetting. To empirically verify this, we show that gradient of intermediate of modularized unleaning model after $S$ iteration is much spread out as compared to normal fine-tuning model's intermediate.

\subsection{Instrumentation of Unlearning}
For an unlearned model, if its distance in parameters space from original model $\mathcal{M}$ is not large, then `roughly' the gradient of loss of unlearning model over retain dataset is orthogonal to that of forget dataset , i.e. the associated dot product is zero. We utilize this proposition in metricizing our unlearning scheme's performance as \textbf{unlearning metric}, by computing cosine similarity between the corresponding gradients, and subtracting it from 1, i.e $1-\frac{\nabla_{\theta}{\mathcal{L}(\mathcal{D}_R)}\cdot\nabla_{\theta}{\mathcal{L}(\mathcal{D}_F)}}{\|\nabla_{\theta}{\mathcal{L}(\mathcal{D}_R)}\|_2\|\nabla_{\theta}{\mathcal{L}(\mathcal{D}_F)}\|_2}$. We compute it as percentage for sake of interpretation.
\\
Overfitting can be seen as the divergence of loss of model over unseen data, while loss of the model on training data is very small. We develope a heuristic metric, which we call \textbf{overfitting metric}, to analyze the overfitting in model from training data perspective as $|(\mathcal{L}(D,\theta))-\text{mean}(|\nabla_{\theta}\mathcal{L}(D,\theta)|)|\in\mathbb{R}$, where $D$ is some input-output pair in $\mathcal{D}$, and $\nabla_{\theta}$ is the gradient with respect to parameters $\theta$. Smaller value of this metric would imply higher degree to which model is overfitted on data $D$.
\\
We also propose a white-box \textbf{model inversion attack}, to visualize the information of training dataset $\mathcal{D}$ from unlearned model, by simply employing algorithm-3, and the $K$ condensed images per class depict the reconstructions of training images per class from unlearned model.

\subsection{Applications of Unlearning}
We propose two new applications for our proposed unlearning schemes.
\subsubsection{Defense Against Membership Inference Attack}
Membership inference attack has been largely linked to overfitting of model on training dataset \cite{shokri2017membership,yeom2018privacy}. Inspired by this, we connect unlearning as a tool to improve membership inference defense, by unlearning to some extent over training data subset, that is more overfitted than remaining. To this extent, we compute overfitting metric over the whole dataset, and perform Otsu binarization over the values of overfitting metric to find the subset of $\mathcal{D}$ that are relatively more overfitted. Then we perform unlearning over few epochs on the detected overfitting samples to achieve defense against membership inference attack.

% \usepackage{multirow}
% \begin{landscape}
\begin{table*}
\centering
\tiny
\begin{tabular}{|c|cccc|cccc|cccc|cccc|c|}
\hline
\multirow{2}{*}{M} & \multicolumn{1}{c|}{RA}    & \multicolumn{1}{c|}{FA}    & \multicolumn{1}{c|}{MIA}   & UT   & \multicolumn{1}{c|}{RA} & \multicolumn{1}{c|}{FA} & \multicolumn{1}{c|}{MIA} & UT & \multicolumn{1}{c|}{RA} & \multicolumn{1}{c|}{FA} & \multicolumn{1}{c|}{MIA} & UT & \multicolumn{1}{c|}{RA} & \multicolumn{1}{c|}{FA} & \multicolumn{1}{c|}{MIA} & UT & \multirow{2}{*}{RBE} \\ \cline{2-17}
                   & \multicolumn{4}{c|}{CIFAR10+MLP}                                                             & \multicolumn{4}{c|}{CIFAR10+CNN}                                                  & \multicolumn{4}{c|}{CIFAR10+ResNet18}                                                & \multicolumn{4}{c|}{CIFAR10+VGG16}                                             &                     \\ \hline
\textbf{R}                  & \multicolumn{1}{c|}{93.56} & \multicolumn{1}{c|}{\textcolor{blue!30}{50.40}} & \multicolumn{1}{c|}{\textcolor{blue!30}{49.54}} & 16.51 & \multicolumn{1}{c|}{96.12}   & \multicolumn{1}{c|}{75.31}   & \multicolumn{1}{c|}{\textcolor{blue!30}{51.04}}    &   43.07 & \multicolumn{1}{c|}{98.87}   & \multicolumn{1}{c|}{82.50}   & \multicolumn{1}{c|}{\textcolor{blue!30}{50.21}}    & 241.31   & \multicolumn{1}{c|}{98.34}   & \multicolumn{1}{c|}{77.54}   & \multicolumn{1}{c|}{\textcolor{blue!30}{50.14}}    &  138.29 & \colorbox{blue1}{0.63}               \\ \hline
\textbf{CF}                 & \multicolumn{1}{c|}{\textcolor{blue!30}{96.85}} & \multicolumn{1}{c|}{77.90} & \multicolumn{1}{c|}{63.01} & 16.56 & \multicolumn{1}{c|}{98.45}   & \multicolumn{1}{c|}{90.24}   & \multicolumn{1}{c|}{56.80}    &  41.91  & \multicolumn{1}{c|}{99.20}   & \multicolumn{1}{c|}{84.96}   & \multicolumn{1}{c|}{50.96}    &   240.63  & \multicolumn{1}{c|}{98.80}   & \multicolumn{1}{c|}{88.12}   & \multicolumn{1}{c|}{54.15}    &  138.38  & \colorbox{blue5}{0.88}              \\ \hline
\textbf{D}                  & \multicolumn{1}{c|}{95.99} & \multicolumn{1}{c|}{55.00} & \multicolumn{1}{c|}{51.42} & 20.61 & \multicolumn{1}{c|}{98.91}   & \multicolumn{1}{c|}{79.51}   & \multicolumn{1}{c|}{50.66}    &  50.00  & \multicolumn{1}{c|}{\textcolor{blue!30}{100.00}}   & \multicolumn{1}{c|}{86.98}   & \multicolumn{1}{c|}{51.79}    &  310.37  & \multicolumn{1}{c|}{89.48}   & \multicolumn{1}{c|}{73.00}   & \multicolumn{1}{c|}{49.46}    & 174.88   & \colorbox{blue4}{0.87}               \\ \hline
\textbf{BD}                 & \multicolumn{1}{c|}{76.64} & \multicolumn{1}{c|}{35.76} & \multicolumn{1}{c|}{53.61} & \textcolor{blue!30}{8.37}  & \multicolumn{1}{c|}{93.59}   & \multicolumn{1}{c|}{\textcolor{blue!30}{63.00}}   & \multicolumn{1}{c|}{73.43}    &   \textcolor{blue!30}{21.16} & \multicolumn{1}{c|}{70.35}   & \multicolumn{1}{c|}{\textcolor{blue!30}{15.02}}   & \multicolumn{1}{c|}{76.08}    &  \textcolor{blue!30}{154.34}  & \multicolumn{1}{c|}{86.67}   & \multicolumn{1}{c|}{\textcolor{blue!30}{25.04}}   & \multicolumn{1}{c|}{77.76}    &  \textcolor{blue!30}{87.16}  & \colorbox{blue7}{1.31}             \\ \hline
\textbf{S}                  & \multicolumn{1}{c|}{91.98} & \multicolumn{1}{c|}{61.66} & \multicolumn{1}{c|}{55.59} & 18.52 & \multicolumn{1}{c|}{99.16}   & \multicolumn{1}{c|}{85.57}   & \multicolumn{1}{c|}{54.53}    &  43.11  & \multicolumn{1}{c|}{98.99}   & \multicolumn{1}{c|}{84.90}   & \multicolumn{1}{c|}{49.81}    &   243.34 & \multicolumn{1}{c|}{\textcolor{blue!30}{98.84}}   & \multicolumn{1}{c|}{84.04}   & \multicolumn{1}{c|}{51.75}    &  147.79  & \colorbox{blue2}{0.78}            \\ \hline
\textbf{P+U}              & \multicolumn{1}{c|}{75.61} & \multicolumn{1}{c|}{52.86} & \multicolumn{1}{c|}{51.71} & 18.01 & \multicolumn{1}{c|}{\textcolor{blue!30}{100.00}}   & \multicolumn{1}{c|}{83.77}   & \multicolumn{1}{c|}{54.01}    &  46.61& \multicolumn{1}{c|}{89.03}   & \multicolumn{1}{c|}{76.88}   & \multicolumn{1}{c|}{50.46}    &   241.91  & \multicolumn{1}{c|}{99.54}   & \multicolumn{1}{c|}{91.14}   & \multicolumn{1}{c|}{55.94}    &  146.41  & \colorbox{blue6}{0.97}               \\ \hline
\rowcolor{lightgray} % Apply the light gray color to this row
\textbf{MU}       & \multicolumn{1}{c|}{89.49} & \multicolumn{1}{c|}{60.70} & \multicolumn{1}{c|}{55.37} & 14.92 & \multicolumn{1}{c|}{91.72}   & \multicolumn{1}{c|}{85.93}   & \multicolumn{1}{c|}{56.85}    &   31.31 & \multicolumn{1}{c|}{94.84}   & \multicolumn{1}{c|}{86.02}   & \multicolumn{1}{c|}{53.20}    &  169.90  & \multicolumn{1}{c|}{90.41}   & \multicolumn{1}{c|}{81.82}   & \multicolumn{1}{c|}{54.44}    &  90.95  &  \colorbox{blue3}{0.80}                 \\ \hline
                   & \multicolumn{4}{c|}{SVHN+MLP}                                                                & \multicolumn{4}{c|}{SVHN+CNN}                                                     & \multicolumn{4}{c|}{SVHN+ResNet18}                                                   & \multicolumn{4}{c|}{SVHN+VGG16}                                                &                     \\ \hline
\textbf{R}                  & \multicolumn{1}{c|}{93.52}      & \multicolumn{1}{c|}{81.20}      & \multicolumn{1}{c|}{51.67}      &     14.82   & \multicolumn{1}{c|}{99.87}   & \multicolumn{1}{c|}{91.88}   & \multicolumn{1}{c|}{50.48}    &  24.88  & \multicolumn{1}{c|}{99.46}   & \multicolumn{1}{c|}{92.20}   & \multicolumn{1}{c|}{49.83}    &   215.69  & \multicolumn{1}{c|}{99.56}   & \multicolumn{1}{c|}{90.73}   & \multicolumn{1}{c|}{\textcolor{blue!30}{50.10}}    &   124.85 &  \colorbox{blue2}{0.61}                 \\ \hline
\textbf{CF}                 & \multicolumn{1}{c|}{\textcolor{blue!30}{96.65}}      & \multicolumn{1}{c|}{87.68}      & \multicolumn{1}{c|}{55.57}      &    15.12   & \multicolumn{1}{c|}{99.81}   & \multicolumn{1}{c|}{96.66}   & \multicolumn{1}{c|}{53.13}    &   25.11  & \multicolumn{1}{c|}{99.74}   & \multicolumn{1}{c|}{93.75}   & \multicolumn{1}{c|}{\textcolor{blue!30}{50.71}}    &  216.07  & \multicolumn{1}{c|}{\textcolor{blue!30}{99.73}}   & \multicolumn{1}{c|}{96.73}   & \multicolumn{1}{c|}{52.60}    &  124.66  &    \colorbox{blue4}{0.69}              \\ \hline
\textbf{D}                  & \multicolumn{1}{c|}{95.03}      & \multicolumn{1}{c|}{83.15}      & \multicolumn{1}{c|}{51.12}      &   18.62    & \multicolumn{1}{c|}{99.77}   & \multicolumn{1}{c|}{92.62}   & \multicolumn{1}{c|}{\textcolor{blue!30}{50.36}}    &   37.63  & \multicolumn{1}{c|}{\textcolor{blue!30}{99.99}}   & \multicolumn{1}{c|}{95.37}   & \multicolumn{1}{c|}{51.37}    &  279.18  & \multicolumn{1}{c|}{99.68}   & \multicolumn{1}{c|}{91.26}   & \multicolumn{1}{c|}{50.22}    &   157.71 &  \colorbox{blue6}{0.83}                \\ \hline
\textbf{BD}                 & \multicolumn{1}{c|}{83.59}      & \multicolumn{1}{c|}{\textcolor{blue!30}{69.68}}      & \multicolumn{1}{c|}{48.85}      &   \textcolor{blue!30}{7.47}    & \multicolumn{1}{c|}{95.91}   & \multicolumn{1}{c|}{\textcolor{blue!30}{60.75}}   & \multicolumn{1}{c|}{87.63}    &  \textcolor{blue!30}{18.80}  & \multicolumn{1}{c|}{91.80}   & \multicolumn{1}{c|}{\textcolor{blue!30}{17.53}}   & \multicolumn{1}{c|}{90.15}    &   \textcolor{blue!30}{139.01}  & \multicolumn{1}{c|}{97.94}   & \multicolumn{1}{c|}{\textcolor{blue!30}{27.64}}   & \multicolumn{1}{c|}{91.90}    &  78.85 & \colorbox{blue5}{0.82}                   \\ \hline
\textbf{S}                  & \multicolumn{1}{c|}{93.45}      & \multicolumn{1}{c|}{83.88}      & \multicolumn{1}{c|}{\textcolor{blue!30}{50.87}}      &   16.59    & \multicolumn{1}{c|}{99.44}   & \multicolumn{1}{c|}{93.04}   & \multicolumn{1}{c|}{51.10}    &  28.63  & \multicolumn{1}{c|}{99.90}   & \multicolumn{1}{c|}{93.73}   & \multicolumn{1}{c|}{49.97}    &  218.90  & \multicolumn{1}{c|}{99.75}   & \multicolumn{1}{c|}{93.35}   & \multicolumn{1}{c|}{50.92}    &  131.46  &  \colorbox{blue3}{0.66}                   \\ \hline
\textbf{P+U}               & \multicolumn{1}{c|}{92.07}      & \multicolumn{1}{c|}{85.11}      & \multicolumn{1}{c|}{51.61}      &    16.34  & \multicolumn{1}{c|}{\textcolor{blue!30}{100.00}}   & \multicolumn{1}{c|}{97.80}   & \multicolumn{1}{c|}{53.78}    &  27.70  & \multicolumn{1}{c|}{69.66}   & \multicolumn{1}{c|}{65.13}   & \multicolumn{1}{c|}{51.33}    &  219.35  & \multicolumn{1}{c|}{19.78}   & \multicolumn{1}{c|}{18.68}   & \multicolumn{1}{c|}{49.88}    &  134.11  &  \colorbox{blue7}{0.88}                \\ \hline
\rowcolor{lightgray} % Apply the light gray color to this row
\textbf{MU}                 & \multicolumn{1}{c|}{93.32}      & \multicolumn{1}{c|}{84.51}      & \multicolumn{1}{c|}{52.87}      &  14.08    & \multicolumn{1}{c|}{97.30}   & \multicolumn{1}{c|}{94.86}   & \multicolumn{1}{c|}{52.86}    &  23.81 & \multicolumn{1}{c|}{97.72}   & \multicolumn{1}{c|}{95.00}   & \multicolumn{1}{c|}{51.74}    &  148.69  & \multicolumn{1}{c|}{94.13}   & \multicolumn{1}{c|}{88.97}   & \multicolumn{1}{c|}{50.44}    &   \textcolor{blue!30}{74.19} &   \colorbox{blue1}{0.48}                \\ \hline
\end{tabular}
%     \caption{Benchmark of reference unlearning algorithms and ours in case of random forgetting of images with forget ratio as 0.1 of total dataset}
%     \label{tab:your_label_here}
% \end{table*}
% % \end{landscape}

% \subsection{Class-wise Forgetting}

% \begin{table*}
% \centering
% \tiny
\vspace{0.1cm} % Space between the two tables
\begin{tabular}{|c|cccc|cccc|cccc|cccc|c|}
\hline
\multirow{2}{*}{M} & \multicolumn{1}{c|}{RA}    & \multicolumn{1}{c|}{FA}    & \multicolumn{1}{c|}{MIA}   & UT   & \multicolumn{1}{c|}{RA} & \multicolumn{1}{c|}{FA} & \multicolumn{1}{c|}{MIA} & UT & \multicolumn{1}{c|}{RA} & \multicolumn{1}{c|}{FA} & \multicolumn{1}{c|}{MIA} & UT & \multicolumn{1}{c|}{RA} & \multicolumn{1}{c|}{FA} & \multicolumn{1}{c|}{MIA} & UT & \multirow{2}{*}{RBE} \\ \cline{2-17}
                   & \multicolumn{4}{c|}{CIFAR10+MLP}                                                             & \multicolumn{4}{c|}{CIFAR10+CNN}                                                  & \multicolumn{4}{c|}{CIFAR10+ResNet18}                                                & \multicolumn{4}{c|}{CIFAR10+VGG16}                                             &                     \\ \hline
\textbf{R}                  & \multicolumn{1}{c|}{86.62} & \multicolumn{1}{c|}{\textcolor{blue!30}{0.00}} & \multicolumn{1}{c|}{91.34} & 17.29 & \multicolumn{1}{c|}{92.76}   & \multicolumn{1}{c|}{\textcolor{blue!30}{0.00}}   & \multicolumn{1}{c|}{94.00}    &   17.38 & \multicolumn{1}{c|}{95.34}   & \multicolumn{1}{c|}{\textcolor{blue!30}{0.00}}   & \multicolumn{1}{c|}{94.40}    & 80.46   & \multicolumn{1}{c|}{90.64}   & \multicolumn{1}{c|}{\textcolor{blue!30}{0.00}}   & \multicolumn{1}{c|}{93.57}    &  46.06 & \colorbox{blue2}{0.62}              \\ \hline
\textbf{CF}                 & \multicolumn{1}{c|}{\textcolor{blue!30}{96.86}} & \multicolumn{1}{c|}{0.02} & \multicolumn{1}{c|}{84.22} & 10.19 & \multicolumn{1}{c|}{\textcolor{blue!30}{99.39}}   & \multicolumn{1}{c|}{0.98}   & \multicolumn{1}{c|}{83.96}    &  18.32  & \multicolumn{1}{c|}{97.61}   & \multicolumn{1}{c|}{\textcolor{blue!30}{0.00}}   & \multicolumn{1}{c|}{93.10}    &   80.43  & \multicolumn{1}{c|}{65.88}   & \multicolumn{1}{c|}{\textcolor{blue!30}{0.00}}   & \multicolumn{1}{c|}{92.26}    &  46.11  & \colorbox{blue4}{0.69}               \\ \hline
\textbf{D}                  & \multicolumn{1}{c|}{79.81} & \multicolumn{1}{c|}{0.06} & \multicolumn{1}{c|}{79.74} & 7.58 & \multicolumn{1}{c|}{94.06}   & \multicolumn{1}{c|}{0.02}   & \multicolumn{1}{c|}{84.32}    &  22.06  & \multicolumn{1}{c|}{\textcolor{blue!30}{99.99}}   & \multicolumn{1}{c|}{9.12}   & \multicolumn{1}{c|}{84.39}    &  103.71  & \multicolumn{1}{c|}{90.35}   & \multicolumn{1}{c|}{\textcolor{blue!30}{0.00}}   & \multicolumn{1}{c|}{92.19}    & 58.45   & \colorbox{blue6}{0.83}             \\ \hline
\textbf{BD}                 & \multicolumn{1}{c|}{87.29} & \multicolumn{1}{c|}{25.08} & \multicolumn{1}{c|}{55.52} & \textcolor{blue!30}{2.77}  & \multicolumn{1}{c|}{98.91}   & \multicolumn{1}{c|}{1.24}   & \multicolumn{1}{c|}{88.16}    &   9.92 & \multicolumn{1}{c|}{80.89}   & \multicolumn{1}{c|}{0.12}   & \multicolumn{1}{c|}{89.96}    &  51.63  & \multicolumn{1}{c|}{97.48}   & \multicolumn{1}{c|}{24.62}   & \multicolumn{1}{c|}{88.29}    &  29.24  & \colorbox{blue5}{0.82}            \\ \hline
\textbf{S}                  & \multicolumn{1}{c|}{95.28} & \multicolumn{1}{c|}{0.02} & \multicolumn{1}{c|}{83.21} & 10.06 & \multicolumn{1}{c|}{99.34}   & \multicolumn{1}{c|}{3.24}   & \multicolumn{1}{c|}{79.46}    &  18.89  & \multicolumn{1}{c|}{96.04}   & \multicolumn{1}{c|}{\textcolor{blue!30}{0.00}}   & \multicolumn{1}{c|}{92.56}    &   81.25 & \multicolumn{1}{c|}{97.23}   & \multicolumn{1}{c|}{\textcolor{blue!30}{0.00}}   & \multicolumn{1}{c|}{90.22}    &  48.69  & \colorbox{blue3}{0.66}             \\ \hline
\textbf{P+U}              & \multicolumn{1}{c|}{71.71} & \multicolumn{1}{c|}{\textcolor{blue!30}{0.00}} & \multicolumn{1}{c|}{92.33} & 9.90 & \multicolumn{1}{c|}{94.95}   & \multicolumn{1}{c|}{\textcolor{blue!30}{0.00}}   & \multicolumn{1}{c|}{92.98}    &  18.75  & \multicolumn{1}{c|}{79.68}   & \multicolumn{1}{c|}{\textcolor{blue!30}{0.00}}   & \multicolumn{1}{c|}{94.67}    &   83.10  & \multicolumn{1}{c|}{\textcolor{blue!30}{99.68}}   & \multicolumn{1}{c|}{\textcolor{blue!30}{0.00}}   & \multicolumn{1}{c|}{88.22}    &  50.81  & \colorbox{blue7}{0.87}               \\ \hline
\rowcolor{lightgray} % Apply the light gray color to this row
\textbf{MU}       & \multicolumn{1}{c|}{88.42} & \multicolumn{1}{c|}{36.76} & \multicolumn{1}{c|}{\textcolor{blue!30}{49.19}} & 4.07 & \multicolumn{1}{c|}{91.51}   & \multicolumn{1}{c|}{64.78}   & \multicolumn{1}{c|}{\textcolor{blue!30}{47.31}}    &   \textcolor{blue!30}{4.77} & \multicolumn{1}{c|}{92.23}   & \multicolumn{1}{c|}{82.66}   & \multicolumn{1}{c|}{\textcolor{blue!30}{52.52}}    &  \textcolor{blue!30}{8.79}  & \multicolumn{1}{c|}{94.74}   & \multicolumn{1}{c|}{91.96}   & \multicolumn{1}{c|}{\textcolor{blue!30}{57.05}}    &  \textcolor{blue!30}{5.29}  &  \colorbox{blue1}{0.47}                  \\ \hline
                   & \multicolumn{4}{c|}{SVHN+MLP}                                                                & \multicolumn{4}{c|}{SVHN+CNN}                                                     & \multicolumn{4}{c|}{SVHN+ResNet18}                                                   & \multicolumn{4}{c|}{SVHN+VGG16}                                                &                     \\ \hline
\textbf{R}                  & \multicolumn{1}{c|}{87.17}      & \multicolumn{1}{c|}{\textcolor{blue!30}{0.00}}      & \multicolumn{1}{c|}{90.26}      &     5.17  & \multicolumn{1}{c|}{99.31}   & \multicolumn{1}{c|}{\textcolor{blue!30}{0.00}}   & \multicolumn{1}{c|}{92.72}    &  8.03  & \multicolumn{1}{c|}{97.88}   & \multicolumn{1}{c|}{\textcolor{blue!30}{0.00}}   & \multicolumn{1}{c|}{93.31}    &   72.00  & \multicolumn{1}{c|}{97.77}   & \multicolumn{1}{c|}{\textcolor{blue!30}{0.00}}   & \multicolumn{1}{c|}{92.95}    &   41.56 &  \colorbox{blue6}{1.32}                 \\ \hline
\textbf{CF}                 & \multicolumn{1}{c|}{\textcolor{blue!30}{96.59}}      & \multicolumn{1}{c|}{0.02}      & \multicolumn{1}{c|}{86.2}      &    5.15   & \multicolumn{1}{c|}{99.72}   & \multicolumn{1}{c|}{7.51}   & \multicolumn{1}{c|}{79.88}    &   7.98  & \multicolumn{1}{c|}{98.02}   & \multicolumn{1}{c|}{\textcolor{blue!30}{0.00}}   & \multicolumn{1}{c|}{93.92}    &  72.08  & \multicolumn{1}{c|}{\textcolor{blue!30}{99.76}}   & \multicolumn{1}{c|}{\textcolor{blue!30}{0.00}}   & \multicolumn{1}{c|}{85.34}    &  41.43  &   \colorbox{blue3}{1.15}               \\ \hline
\textbf{D}                  & \multicolumn{1}{c|}{88.91}      & \multicolumn{1}{c|}{12.46}      & \multicolumn{1}{c|}{73.05}      &   6.31    & \multicolumn{1}{c|}{98.91}   & \multicolumn{1}{c|}{1.55}   & \multicolumn{1}{c|}{85.25}    &   10.16  & \multicolumn{1}{c|}{\textcolor{blue!30}{99.99}}   & \multicolumn{1}{c|}{53.00}   & \multicolumn{1}{c|}{73.16}    &  93.17  & \multicolumn{1}{c|}{98.36}   & \multicolumn{1}{c|}{\textcolor{blue!30}{0.00}}   & \multicolumn{1}{c|}{91.21}    &  52.66 &    \colorbox{blue5}{1.24}               \\ \hline
\textbf{BD}                 & \multicolumn{1}{c|}{91.72}      & \multicolumn{1}{c|}{14.02}      & \multicolumn{1}{c|}{78.23}      &   \textcolor{blue!30}{2.63}    & \multicolumn{1}{c|}{95.91}   & \multicolumn{1}{c|}{1.42}   & \multicolumn{1}{c|}{91.02}    &  4.97  & \multicolumn{1}{c|}{94.51}   & \multicolumn{1}{c|}{7.40}   & \multicolumn{1}{c|}{91.75}    &   46.49  & \multicolumn{1}{c|}{99.69}   & \multicolumn{1}{c|}{3.24}   & \multicolumn{1}{c|}{91.28}    &  26.37 & \colorbox{blue2}{1.13}                   \\ \hline
\textbf{S}                  & \multicolumn{1}{c|}{94.20}      & \multicolumn{1}{c|}{0.08}      & \multicolumn{1}{c|}{86.55}      &   5.95    & \multicolumn{1}{c|}{99.63}   & \multicolumn{1}{c|}{27.53}   & \multicolumn{1}{c|}{72.40}    &  9.16  & \multicolumn{1}{c|}{98.75}   & \multicolumn{1}{c|}{\textcolor{blue!30}{0.00}}   & \multicolumn{1}{c|}{94.01}    &  72.97  & \multicolumn{1}{c|}{99.80}   & \multicolumn{1}{c|}{\textcolor{blue!30}{0.00}}   & \multicolumn{1}{c|}{88.27}    &  43.63  & \colorbox{blue4}{1.17}                     \\ \hline
\textbf{P+U}               & \multicolumn{1}{c|}{88.85}      & \multicolumn{1}{c|}{0.02}      & \multicolumn{1}{c|}{89.38}      &    6.06  & \multicolumn{1}{c|}{\textcolor{blue!30}{100.00}}   & \multicolumn{1}{c|}{38.28}   & \multicolumn{1}{c|}{69.54}    &  9.07  & \multicolumn{1}{c|}{95.08}   & \multicolumn{1}{c|}{\textcolor{blue!30}{0.00}}   & \multicolumn{1}{c|}{94.05}    &  74.15  & \multicolumn{1}{c|}{47.58}   & \multicolumn{1}{c|}{\textcolor{blue!30}{0.00}}   & \multicolumn{1}{c|}{78.02}    &  45.26  &  \colorbox{blue7}{1.43}                 \\ \hline
\rowcolor{lightgray} % Apply the light gray color to this row
\textbf{MU}                 & \multicolumn{1}{c|}{92.36}      & \multicolumn{1}{c|}{66.80}      & \multicolumn{1}{c|}{\textcolor{blue!30}{55.21}}      &  2.73    & \multicolumn{1}{c|}{94.86}   & \multicolumn{1}{c|}{90.48}   & \multicolumn{1}{c|}{\textcolor{blue!30}{51.50}}    &  \textcolor{blue!30}{3.85} & \multicolumn{1}{c|}{98.93}   & \multicolumn{1}{c|}{98.66}   & \multicolumn{1}{c|}{\textcolor{blue!30}{55.26}}    &  \textcolor{blue!30}{8.66}  & \multicolumn{1}{c|}{93.71}   & \multicolumn{1}{c|}{80.11}   & \multicolumn{1}{c|}{\textcolor{blue!30}{47.45}}    &   \textcolor{blue!30}{5.27} &                  \colorbox{blue1}{0.58}   \\ \hline
\end{tabular}
    \caption{Benchmark of reference unlearning algorithms and ours in case of random image forgetting (first table) the 10 percent of total training dataset and class forgetting  of images (second table)}
    \label{tab:your_label_here}
\end{table*}
\subsubsection{Unlearning in Dataset Condensation}
One of the characteristics of modular unlearning is in the flexibility of changing the architecture of parameters of \textbf{final}, and retraining on few remembrance samples can allow quick regain of accuracy. In light of this, we propose a route of new dataset condensation strategy that results in a `condensed model', rather than images, as condensed representation. More precisely, in our modular training's offline part, we form an autoencoder topology on the \textbf{beginning} and \textbf{intermediate}, i.e. the input and output dimensions of \textbf{beginning}-\textbf{intermediate} composition is same. The output of the this structure over the remembrance samples can be used a substitute for condensed images. Hence, the condensed model can be combined with any new deep learning architecture (which would serve the role of \textbf{final}), and it can be quickly trained over the remembrance samples to gain accuracy over the original dataset. The reason for this route of dataset condensation is that this strategy of dataset condensation allows unlearning, which is not feasible for image-driven dataset condensation, while at the same time being very fast and accurate, but at the cost of increasing the parameter count of any new deep learning model it is applied to. Simply by performing the offline and online phase of our unlearning scheme with the assumptions of \textbf{beginning}, \textbf{intermediate} and \textbf{final}, and at the end replacement of \textbf{final} with a new deep learning model leads to augmented model which when trained on remembrance samples rapidly leads to model that is approximately equivalent to same model if it was trained only on retain dataset.

%%%%%%%%%%%%%%%%%%%%%%%%%%%%%%%%%%%%%%%%%%%%%%%%%%%%%%%%%%%%%%%%%%%%%%%%%%%%%%%%%%%%%%%%%%%%%%%%%
\begin{figure}[t]
\centering
\hspace*{-0.1in}
\includegraphics[width=8.5cm, height=3.0cm]{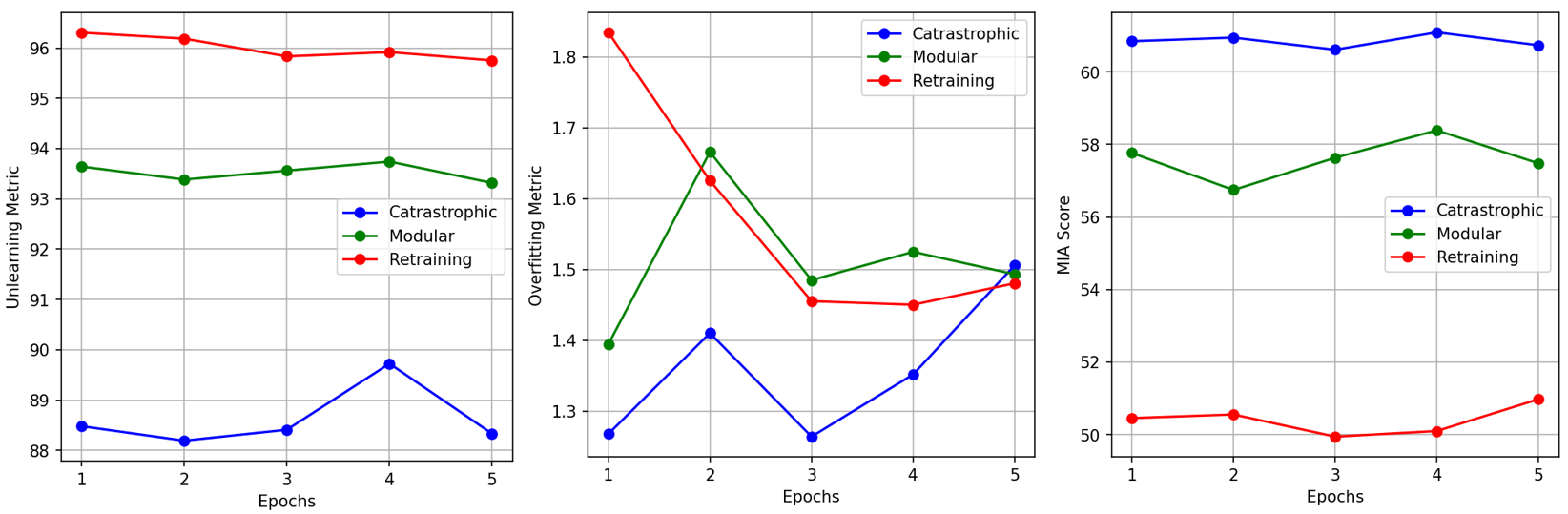}
\caption{Evolution of UM, OM and MIA for first first epochs of modular unlearning and catastrophic forgetting over VGG16 on CIFAR10}
\end{figure}
%%%%%%%%%%%%%%%%%%%%%%%%%%%%%%%%%%%%%%%%%%%%%%%%%%%%%%%%%%%%%%%%%%%%%%%%%%%%%%%%%%%%%%%%%%%%%%%

\section{Performance Evaluation}

\subsection{Experimental Settings}

\subsubsection{Datasets}
We conduct experiments over CIFAR-10 and SVHN, as in 
\cite{bourtoule2021machine} to evaluate unlearning performance as well as developing applications in image classification task.

\subsubsection{Unlearning Baselines}
We implement following approximate unlearning baselines for the unlearning performance comparison. First one is  \textbf{Retraining (R)}, where we train the randomly initialized model on the retain dataset, as the naive unlearning method. In \textbf{Catastrophic Forgetting (CF)}, we train the pre-trained model on retain dataset. Inspired by \cite{kurmanji2023towards}, we implement \textbf{Distillation (D)} based unlearning only focus on the distilling the given model on retain dataset, without the increasing the KL-divergence over forget dataset, which then leads to increase to MIA score, which then is compensated by `Rewinding' procedure. This is sufficient as we are exploring applications discussed in \cite{kurmanji2023towards}, where negative KL-divergences are needed. Another distillation based unlearning methodology we implement is\textbf{Bad Teacher based Distillation (BD)} \cite{chundawat2023can}, where we utilize competent (pretrained model) and incompetent (randomly initialized model) teachers to minimize the weighted KL-divergence between student and the two teachers, over the forget dataset, and randomly sampled retain dataset. In \textbf{Sparisity Regularized Unlearning (S)} \cite{jia2023model} we basically perform catastrophic forgetting with regularization loss by $\|\theta\|_1$, where $\theta$ is the vector containing parameters of the neural network \cite{jia2023model}. We also adopt \textbf{Pruning and then unlearning (P+U)} \cite{jia2023model}, we perform model prunning via synaptic flow \cite{tanaka2020pruning} on the pretrained model, and then perform unlearning. For the unlearning part, we simply train the resultant prunned model on retain set.

%%%%%%%%%%%%%%%%%%%%%%%%%%%%%%%%%%%%%%%%%%%%%%%%%%%%%%%%%%%%%%%%%%%%%%%%%%%%%%%%%%%%%%%%%%%%%%%%%
\begin{figure*}[t]
\centering
\hspace*{-0.1in}
\includegraphics[width=15.0cm, height=5.0cm]{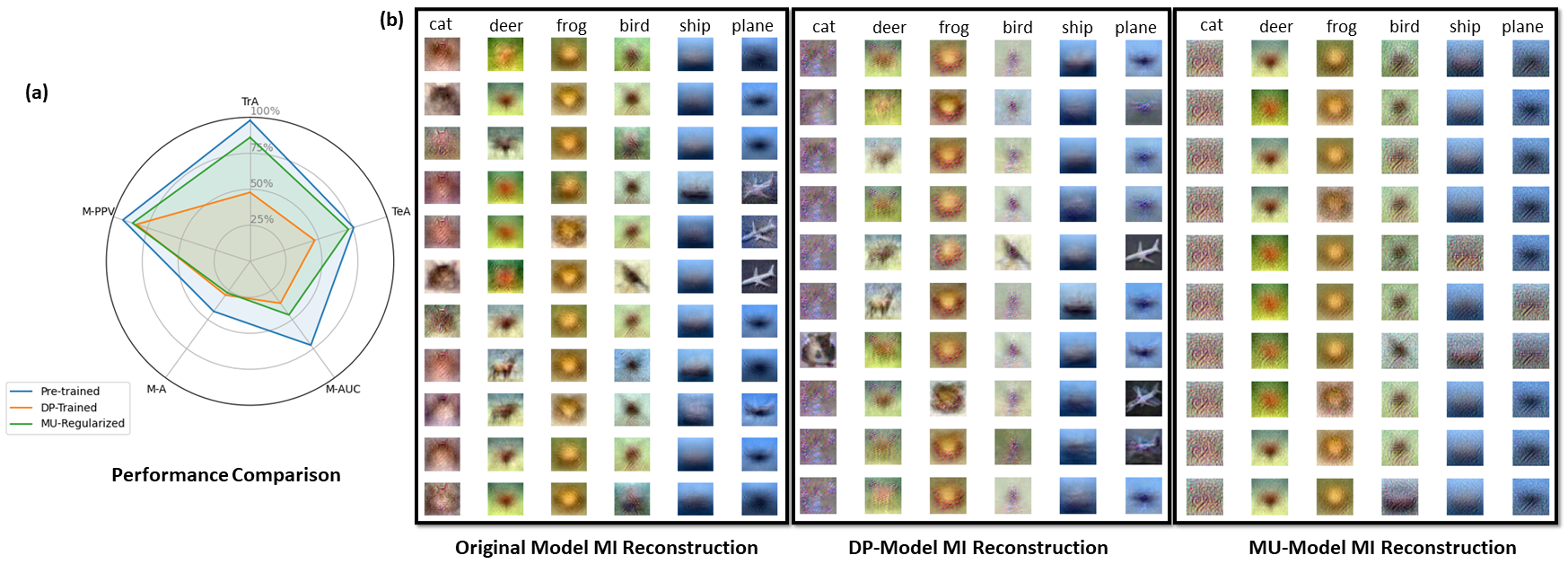}
\caption{Proposed model inversion attack based reconstruction of images per class of CIFAR-10 dataset from original model, model training with differentially-private Adam based optimization and proposed unlearning based regularization of model}
\end{figure*}
%%%%%%%%%%%%%%%%%%%%%%%%%%%%%%%%%%%%%%%%%%%%%%%%%%%%%%%%%%%%%%%%%%%%%%%%%%%%%%%%%%%%%%%%%%%%%%%

%%%%%%%%%%%%%%%%%%%%%%%%%%%%%%%%%%%%%%%%%%%%%%%%%%%%%%%%%%%%%%%%%%%%%%%%%%%%%%%%%%%%%%%%%%%%%%%%%
\begin{figure}[t]
\centering
\hspace*{-0.1in}
\includegraphics[width=7.5cm, height=3.6cm]{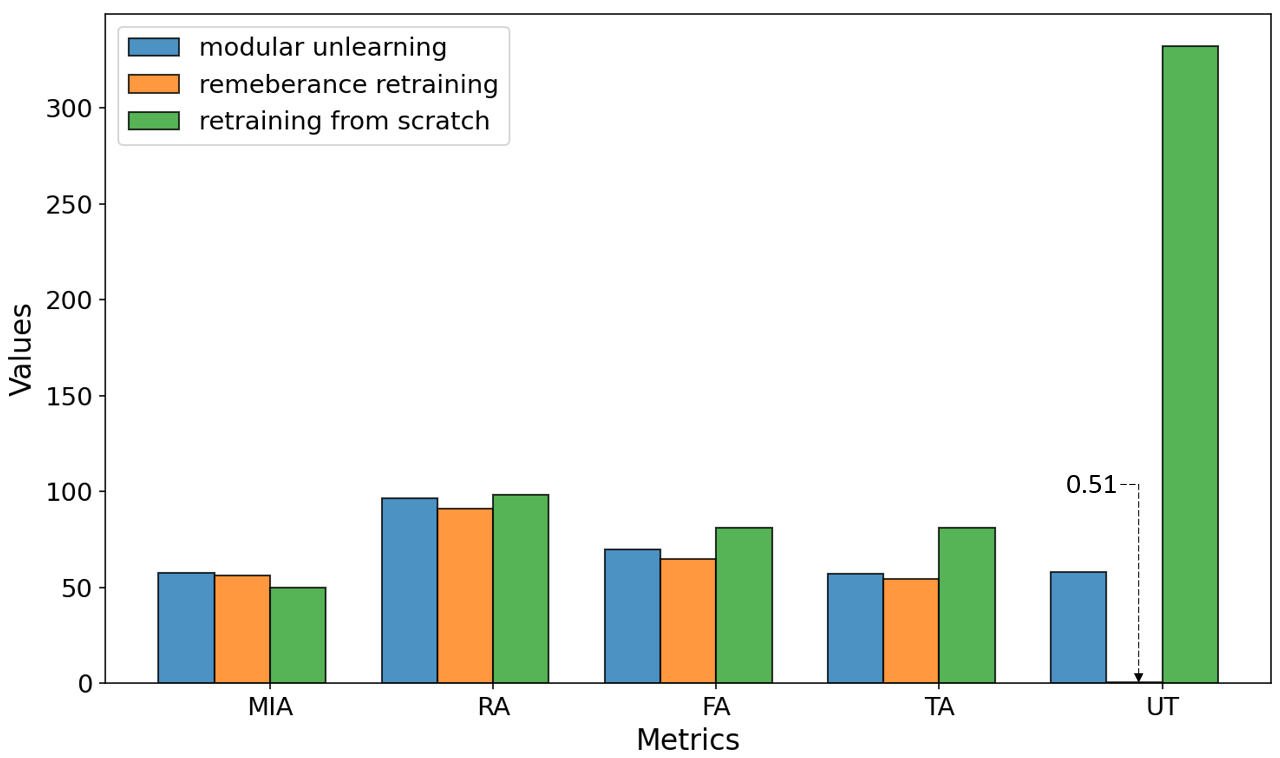}
\caption{Benchmarking of Unlearning in Condensation setting, where the goal is to unlearn the data from condensed knowledge which can be quickly used to train another model}
\end{figure}
%%%%%%%%%%%%%%%%%%%%%%%%%%%%%%%%%%%%%%%%%%%%%%%%%%%%%%%%%%%%%%%%%%%%%%%%%%%%%%%%%%%%%%%%%%%%%%%

\subsubsection{Implementations}
We implemented the baselines and all experimentation in Python 3.10.12, within the framework of Pytorch library v2.1.0, in Windows Subsystem for Linux (WSL2) and GPU hardware as NVIDIA GeForce RTX 3090. For all experiments involving training, we use Adam optimizer with fixed learning rate. The architecture wise hyperparameters of the unlearning models are shown in table-1.

\subsubsection{Metrics}
In order to elucidate on utility from the unlearning procedure, we utilize metrics like accuracy of unlearning model over retain dataset as \textbf{(RA)}, accuracy over forget dataset as \textbf{FA}, total time (in seconds) for unlearning algorithm completion and membership inference attack accuracy as \textbf{MIA} by building a logistic classifier over losses of the unlearning model over forget and test dataset.
\\
In order to rank the unlearning performance on previous all four metrics for a single dataset, by associating weights to \textbf{RA}, \textbf{FA}, \textbf{MIA}, and \textbf{UT} scores as 1, 0.5, 1 and 1 respectively. Here we first calculated absolute difference of the previous metric evaluations from the best metric found for each metric. Then performing min-max normalization of the resultant score over all score evaluations on the same dataset, then apply weighted averaging the RA, FA, MIA and UT scores. Afterwards, averaging the results for each model to getting a relative notion of performance measure, called relative best error, which should be small as possible.
\\
In order the compute the membership inference attack accuracy of model over whole dataset $\mathcal{D}$, we employ shadow model based strategy \cite{shokri2017membership}, where the shadow models are multi-layer perceptrons (MLP), and the final classifier is a logistic regressor. In addition, we also compute training and test dataset accuracy \textbf{TrA} and \textbf{TeA} for specific cases.

\subsection{Balance Between Major Unlearning Metrics}
We present the comparison of metrics \textbf{RA}, \textbf{FA}, \textbf{MIA} and \textbf{UT} over reference unlearning algorithm, and ours in table-1, where we attempt to forget either 10 percent of training dataset in the case of random forgetting or attempting to forget entire class in case of classwise forgetting. It can be observed that while our proposed unlearning methodology does not surpasses in all metrics over other approaches, it finds a good median between all metrics, while other approaches fail to do so. In order to quickly metricize this, we tabulated the RBE values over dataset case, and it can be discovered that our model ranks first in 3 out of 4 case, while ranking third in case of random forgetting in CIFAR-10 dataset. It should be noted that \textbf{BD} achieves lesser unlearning time than ours, especially in random forgetting case, is because it utilizes 30\% randomly sampled retain dataset, while our approach systematically reduces dataset based on value of $K$. For reference, here $K=450$, and average percentage of reduced retain dataset size is 80\% in the case of random forgetting.

\subsection{Relationship between Unlearning Metric and Membership Inference Attack}
We computed the unlearning metric, overfitting metric and \textbf{MIA} for first few epochs of unlearning, in the case of \textbf{R}, \textbf{CF} and \textbf{MU}. The reason for chosing first few epochs is to prevent the unlearned parameters not deviate largely from original parameters. The results are shown in figure 2, where it can be noted that proposed \textbf{MU} consistently performs much better than \textbf{CF}, on which it is based on, approaches the effects of \textbf{R}. It can be noted that unlearning metric is significantly correlated with \textbf{MIA}, while the correlation between overfitting metric and \textbf{MIA} are less visible at this early point in training.

\subsection{Competitor to Differential Privacy}
We attempt to exercise our strategy of defending against membership inference attack, and comparing it with differential privacy based solution. We performed our proposed unlearning based regularization over VGG16 trained on CIFAR10, in comparison with training the VGG16 using DP-Adam \cite{bu2022differentially1,bu2022differentially2}. The results are shown in figure 3. It is abundantly clear that while proposed unlearning based regularization has similar privacy effects as DP-Adam, it has significantly higher utility in close proximity with that of original model.

\subsection{Unlearning in Dataset Condensation}
In order to show the effectiveness of our proposed unlearning in condensation, we created a convolution-deconvolution based autoencoder architecture to be utilized as \textbf{beginning} and \textbf{intermediate} in our strategy, while a MLP is assigned as \textbf{final}. We attempt to forget 10 percent of random images from CIFAR-10 dataset. After completion of offline and online phase of proposed unlearning, we subustitute the \textbf{final} with VGG16, which we intended to train on condensed dataset (not containing forget dataset). The results are vivid in figure 4, where the resultant model is approaches the privacy and utility of \textbf{R}, at significant advantage over the retraining VGG16 on retain dataset in terms of training time.

\section{Conclusion}
In this paper, we proposed a new unlearning scheme through interplay between catastrophic forgetting and dataset condensation. It has shown to be best balanced approximate unlearning scheme in terms of privacy, utility and efficiency through extensive experiments and results. We showed its application in protecting the privacy of deep learning model, as well as unlearning in dataset condensation. We envision our work as stepping stone for further investigation into relationship between unlearning and dataset condensation.

\section*{Acknowledgments}
This work was done in Korea National University of Transportation, Chungju, South Korea. The work was partially funded by the IITP project, titled ``Development of a high-performance visual discovery platform for understanding and predicting real-time large-scale shape data".

%%%%%%%%% REFERENCES
% % {\small
% \bibliographystyle{ieee}
% \bibliography{egbib}
% % }

% % \vfill

\bibliographystyle{ieeetr}

\bibliography{ref.bib}

\appendix

\section{Recap of Mathematical Notation}

\setcounter{equation}{0}
% \label{sec:intro}

We define the original dataset as $\mathcal{D}=\{T_i, l_i\}_{i=1}^{N_D}$,  where $T_i$ is the $i^{th}$ training image and $l_i$ is the $i^{th}$ label. $\mathcal{D}$ can be partitioned into the forget dataset $\mathcal{F}=\{T_{f_i}, l_{f_i}\}_{i=1}^{N_F}$ and retain dataset $\mathcal{R}=\{T_{r_i}, l_{r_i}\}_{i=1}^{N_R}$. Here, $\mathcal{F} \cup \mathcal{R} = \mathcal{D}$, and $\mathcal{F} \cap \mathcal{R} = \emptyset$. For the sake of representation, we represent the images in $\mathcal{D}$ as $T = \bigcup_{i=1}^{c} \bigcup_{j=1}^{n} T_{ij}$, where $c$ is the total number of classes in $\mathcal{D}$, $n$ is the number of images per class such that $N_D=nc$ and $T_{ij}$ represents the image in $i^{th}$ class and $j^{th}$ index.
\\
The forget set $\mathcal{F}$ represents the part of the training dataset $\mathcal{D}$ to be forgotten by a model, trained on $\mathcal{D}$ using a twice differentiable loss function $\mathcal{L}$. Lets call this trained model as $\mathcal{M}_{\theta}$ with trained parameters $\theta$, which is achieved by minimizing $\mathcal{L}(\mathcal{D},\theta)=\sum_{i=1}^{N_D}((\mathcal{M}(T_i),l_i),\theta)$. The goal of dataset reduction framework is reduce $\mathcal{R}\rightarrow{\mathcal{R}_{\text{red}}}$, such that $N_r=|\mathcal{R}_{\text{red}}|$ and $N_r<N_R$. Within the dataset reduction framework, the images of each class are grouped into $K$ clusters, such that $K<n$. Finally, after unlearning procedure, the model $\mathcal{M}$ gains parameters $\theta^{*}$.

\section{Derivation of Unlearning Metric}
\subsection{Relation between Original and Unlearned Parameters}
Suppose there exists a parameter $\zeta\in[0,1]$, and we define a modified loss function as.
\begin{equation} \label{eq1}
\mathcal{L}(\mathcal{D},\theta(\zeta),\zeta)=\mathcal{L}(\mathcal{R},\theta(\zeta))+\zeta{\mathcal{L}(\mathcal{F},\theta(\zeta))}
\end{equation}
such that,
\begin{equation} \label{eq1}
\theta(\zeta)=\theta^{*}+\sum_{i=1}^{\infty}{\zeta^i{\theta_i}}
\end{equation}
As we deform $\zeta$ from 0 to 1, then $\mathcal{L}(\mathcal{D},\theta,\zeta)$ changes from $\mathcal{L}(\mathcal{R},\theta(0))$ to $\mathcal{L}(\mathcal{R},\theta(1))+ \mathcal{L}(\mathcal{F},\theta(1))$, thus representing that $\mathcal{L}(\mathcal{D},\theta(\zeta),\zeta)$ deforms from loss over retain dataset to loss over whole training dataset. Suppose if $\theta(\zeta)$ is local minima of $\mathcal{L}(\mathcal{D},\theta(\zeta),\zeta)$, then.

\begin{equation} \label{eq1}
\nabla_{\theta}\mathcal{L}(\mathcal{R},\theta(\zeta))+\zeta\nabla_{\theta}{\mathcal{L}(\mathcal{F},\theta(\zeta))}=0
\end{equation}
In the lines of perturbation theory \cite{liao2003beyond}, by choosing $\zeta$ small enough, we make first order approximation of $\theta\approx\theta^*+\zeta\theta_1$, such that $\zeta^2\rightarrow{0}$.
After inserting it into Equation (5).

\begin{equation} \label{eq1}
\nabla_{\theta}\mathcal{L}(\mathcal{R},\theta^*+\zeta\theta_1)+\zeta\nabla_{\theta}{\mathcal{L}(\mathcal{F},\theta^*+\zeta\theta_1)}=0
\end{equation}

By performing Taylor approximation around $\theta^*$, we get.
\begin{equation} 
\begin{split}
\nabla_{\theta}\mathcal{L}(\mathcal{R},\theta^*) + \zeta\nabla^2_{\theta}\mathcal{L}(\mathcal{R},\theta^*)\theta_1 + \zeta\nabla_{\theta}\mathcal{L}(\mathcal{F},\theta^*) \\
+ \zeta^2\nabla^2_{\theta}\mathcal{L}(\mathcal{F},\theta^*)\theta_1+o(\zeta\theta_1) = 0
\end{split}
\end{equation}

Omitting $o(\zeta\theta_1)$ term, we balance the coefficients of $1$ and $\zeta$ in Equation (5), since $\zeta$ is arbitrary and independent parameter. Then $\nabla_{\theta}\mathcal{L}(\mathcal{R},\theta^*)=0$, implying that $\theta^*$ is the minima of $\mathcal{L}(\mathcal{R},\theta^*)$. For the case of $\zeta$ coefficients,

\begin{equation} 
\begin{split}
\theta_1=-\nabla^2\mathcal{L}(\mathcal{R},\theta_0)^{-1}\nabla\mathcal{L}(\mathcal{F},\theta_0)
\end{split}
\end{equation}

Using Equations (2) and (6), we achieve first order approximation of $\theta$ as,

\begin{equation}
\boxed{
\theta=\theta^*-\zeta\nabla^2\mathcal{L}(\mathcal{R},\theta^*)^{-1}\nabla\mathcal{L}(\mathcal{F},\theta^*) +o(\zeta)
}
\end{equation}

\textbf{Remark:} 
\begin{itemize}
    \item The Equation (7) resembles the derivation of change of parameters under small influence of a new training sample \cite{koh2017understanding}. Through Cauchy-Schawarz inequality, it can be observed that $\|\theta-\theta^*\|_2\leq\zeta\|\nabla^2\mathcal{L}(\mathcal{R},\theta^*)^{-1}\|_F\|\nabla\mathcal{L}(\mathcal{F},\theta^*)\|_2$, where $\|\|_F$ is the Forbenious norm. Since $\|\nabla^2\mathcal{L}(\mathcal{R},\theta^*)^{-1}\|_F>0$ as due to stationary condition of  $\mathcal{L}(\mathcal{R},\theta^*)$ \cite{richardson2011advanced} and strict positive definite condition \cite{strang2022introduction}, therefore necessarily the performance of unlearned model (optimized from original pretrained model) deteriorates over forget dataset, because $\|\nabla\mathcal{L}(\mathcal{F},\theta^*)\|_2>0$ when $\|\theta-\theta^*\|_2>0$ and $\|\nabla\mathcal{L}(\mathcal{F},\theta^*)\|_2=0$ if and only if $\|\theta-\theta^*\|_2=0$. Thus, Equation (7) is a mathematical statement about catastrophic forgetting.

    \item If instead of defining loss as summation of loss over individual samples, we averaged the individual losses, then we would have to redefine Equation (1) as $\mathcal{L}(\mathcal{D},\theta(\zeta),\zeta)=((1-\zeta)+\zeta\frac{N_R}{N_D})\mathcal{L}(\mathcal{R},\theta(\zeta))+\zeta\frac{N_F}{N_D}{\mathcal{L}(\mathcal{F},\theta(\zeta))}$. Then through same sequence of steps, we would arrive at modified version of Equation (7) as.
\begin{equation*}
\boxed{
\begin{aligned}
\theta &= \theta^* - \zeta \frac{N_F}{N_D} \nabla^2\mathcal{L}(\mathcal{R},\theta^*)^{-1} \nabla\mathcal{L}(\mathcal{F},\theta^*)+ o(\zeta)
\end{aligned}
}
\end{equation*}

We do not progress in this fashion, since first the positions of local minima does not change by scaling the loss function, and secondly the resultant analysis is simpler to deal with, needless to say it does not change the consequential results.
    
\end{itemize}

\par
If we progressively substitute second-order, to $n^{\text{th}}$ order approximation of $\theta$ from Equation (2) into Equation (3), and apply similarly first order Taylor approximation around $\theta_0$, we can derive the expressions for $\theta_2$ up to $\theta_n$, by equating the coefficients of $\zeta^2$ to $\zeta^n$ to zero. For example, $\theta_2$ can be derived with this strategy as.
\begin{equation} 
\begin{split}
\theta_2=\nabla^2\mathcal{L}(\mathcal{F},\theta^*)^{-1}\nabla\mathcal{L}(\mathcal{F},\theta^*)\nabla^2\mathcal{L}(\mathcal{R},\theta^*)^{-1}\nabla\mathcal{L}(\mathcal{F},\theta^*)
\end{split}
\end{equation}

So that substituting Equations (6) and (8) into Equation (2) would give a second order approximation of $\theta$.
% \begin{equation} 

\begin{equation*}
\boxed{
\begin{aligned}
\theta &= \theta^* - \zeta \nabla^2 \mathcal{L}(\mathcal{R}, \theta^*)^{-1} \nabla \mathcal{L}(\mathcal{F}, \theta^*) \nonumber \\
&\quad + \zeta^2 \nabla^2 \mathcal{L}(\mathcal{F}, \theta^*)^{-1} \nabla \mathcal{L}(\mathcal{F}, \theta^*) \nabla^2 \mathcal{L}(\mathcal{R}, \theta^*)^{-1} \\
&\quad \nabla \mathcal{L}(\mathcal{F}, \theta^*) + o(\zeta^2)
\end{aligned}
}
\end{equation*}

% \end{equation}

\textbf{Remark:} We highlight this equation, in light of \cite{koh2017understanding}, where making $\zeta\rightarrow{1}$ besides $\zeta\rightarrow{0}$ for deriving influence function, i.e $\frac{d}{d\zeta}(\theta-\theta^*)|_{\zeta=1}$ can lead to new analytical results with the underlying intuition of equally up-weighting the new data samples (samples in $\mathcal{F}$ in current case), instead of infinitesimal up-weight of new-data samples. This course of study we leave for future work.

\subsection{Persistance of Loss of Unlearned Model on Retain Dataset}

For input-output pair $(T_i,l_i)\in{\mathcal{R}}$, we expand the loss of pretrained model around unlearned parameters as a Taylor's expansion.
\begin{equation} 
\begin{split}
\mathcal{L}((\mathcal{M}(T_i),l_i),\theta)=\mathcal{L}((\mathcal{M}(T_i),l_i),\theta^*)+\\ \nabla\mathcal{L}((\mathcal{M}(T_i),l_i),\theta^*)\cdot\delta\theta+o(\|\delta\theta\|_2)
\end{split}
\end{equation}

where $\delta\theta=\theta-\theta^*$. We wish to the loss of pretrained and unlearned model over retain samples to be retain to conform to unlearning utility principle. Therefore we rewrite Equation (11), and omitting $o(\|\delta\theta\|_2)$ terms.
\begin{equation} 
\begin{split}
\mathcal{L}((\mathcal{M}(T_i),l_i),\theta)-\mathcal{L}((\mathcal{M}(T_i),l_i),\theta^*) =\\ \nabla\mathcal{L}((\mathcal{M}(T_i),l_i),\theta^*)\cdot\delta\theta
\end{split}
\end{equation}

Utilizing Equation (7) with omission of $o(\zeta)$ terms, Equation (11) changes to.
\begin{equation} 
\begin{split}
\mathcal{L}((\mathcal{M}(T_i),l_i),\theta)-\mathcal{L}((\mathcal{M}(T_i),l_i),\theta^*) =\\ \nabla\mathcal{L}((\mathcal{M}(T_i),l_i),\theta^*)\cdot \\(\zeta\nabla^2\mathcal{L}(\mathcal{R},\theta^*)^{-1}\nabla\mathcal{L}(\mathcal{F},\theta^*))
\end{split}
\end{equation}

By summing over all $(T_i,l_i)\in{\mathcal{R}}$ and scaling both sides of equation with $\frac{1}{N_R}$, we write Equation (12) as follows.
\begin{equation} 
\begin{split}
|\mathcal{L}(\mathcal{R},\theta^*)-\mathcal{L}(\mathcal{R},\theta)| =\\ \zeta|\nabla\mathcal{L}(\mathcal{R},\theta^*)\cdot (\nabla^2\mathcal{L}(\mathcal{R},\theta^*)^{-1}\nabla\mathcal{L}(\mathcal{F},\theta^*))|
\end{split}
\end{equation}

\subsubsection{Nature of Inverse Hessian}

The Hessian in Equation (13) is positive definite \cite{richardson2011advanced}, and so is its inverse \cite{horn2012matrix}. The $\nabla^2\mathcal{L}(\mathcal{R},\theta^*)^{-1}$ has positive eigenvalues \cite{strang2022introduction} and can be represented as $\{\frac{1}{\lambda_i}\}_{i=1}^n$, where $\lambda_i$ is the eigenvalue of $\nabla^2\mathcal{L}(\mathcal{R},\theta^*)$, which are also positive. Due to definite of positive definite matrices, $\nabla^2\mathcal{L}(\mathcal{R},\theta^*)^{-1}\nabla\mathcal{L}(\mathcal{F},\theta^*)$ should not change the direction of $\nabla\mathcal{L}(\mathcal{F},\theta^*)$ by more than $\frac{\pi}{2}$ radians, due to positive dot product condition \cite{strang2022introduction}. 
\\It has been pointed out that during optimization, the eigenvalues of Hessian concentrate around zero with roughly equal distribution, with few outliers \cite{ghorbani2019investigation}. Taking this observation as starting point, we want to claim that the eigenvalues of $\nabla^2\mathcal{L}(\mathcal{R},\theta^*)^{-1}$ even more concentrated (small vairance) and uniform. This motivation is based on eigendecomposition \cite{strang2022introduction} of positive definite matrix into $UDU^{-1}$, where $U$ is a unitary matrix, whose columns are eigenvectors of positive definite matrix, forming an orthonormal basis, and $D$ is a diagonal matrix with positive eigenvalues. In this interpretation, the $U^{-1}$ and $U$ move towards and back from positive definite matrix's orthonormal eigenbasis, like an encoder-decoder setup in deep learning. Within the eigenbasis (encoded domain), each component of resultant vector gets scaled with positive eigenvalues. We wish to concentrate and uniform the eigenvalues of $\nabla^2\mathcal{L}(\mathcal{R},\theta^*)$, so that it has dominating function of just scaling the components of $\nabla\mathcal{L}(\mathcal{F},\theta^*)$ in equal amount, i.e. it approximately acts as a scaler to a vector it acts upon. This effect's significance will be pointed out later.
\\

Without loss of generality, we define variance of a finite positive sequence $\{a_i\}_{i=1}^n$ as $V\{a\}=\sum_{i=1}^n{(M\{a\}-a_i)^2}$ and mean as $M\{a\}=\frac{1}{n}\sum_{j=1}^n{a_j}$. Henceforth, we compute variance of eigenvalues of $\nabla^2\mathcal{L}(\mathcal{R},\theta^*)^{-1}$ by starting with obvious first computation step and then applying arithematic mean-harmonic mean inequality on the first term in outer summation,
\begin{equation} 
\begin{split}
V\{\frac{1}{\lambda}\}\leq\sum_{i=1}^n(\frac{1}{\frac{1}{n}\sum_{j=1}^n{\lambda_j}}-\frac{1}{\lambda_i})^2
\end{split}
\end{equation}

\begin{equation} 
\begin{split}
=\sum_{i=1}^n(\frac{\frac{1}{n}\sum_{j=1}^n\lambda_j-\lambda_i}{\frac{\lambda_i}{n}\sum_{j=1}^n\lambda_j})^2
\end{split}
\end{equation}

By apply Cauchy-Schwarz inequality in attempt to splitting sum over numerator and denomenator, and performing trivial computations, we arrive at.
\begin{equation} 
=\frac{V\{\lambda\}}{M\{\lambda\}}\sum_{i=1}^n\frac{1}{\lambda_i}
\end{equation}
Next we reduce the above the harmonic like sum $\sum_{i=1}^n\frac{1}{\lambda_i}$, by applying Abel summation. We chose a sequence $\{b_k\}_{k=1}^n$ such that $b_k=1$, $B(t)=\sum_{0\leq k \leq t}b_k= t+1$, and $\phi(t)=\frac{1}{\lambda_t}$, such that $\phi(t)$ is descending ordered sequence over $\{\frac{1}{\lambda_i}\}_{i=1}^n$, such that $\phi(0)=\frac{1}{\lambda_{\min}}$ and $\phi(n-1)=\frac{1}{\lambda_{\max}}$. We then define the Abel summation formula as.
\begin{equation} 
\begin{split}
\sum_{0\leq i \leq n-1} {b_i\phi(i)}= B(n-1)\phi(n-1)-B(0)\phi(0)-\\ \int_{0}^{n-1}B(z)\phi'(z)dz
\end{split}
\end{equation}
We model the descending nature of $\phi(u)$ via $\phi(u)=\frac{1}{\lambda_{\min}}e^{-ku}$, where $k$ can be derived by substituting $\phi(n-1)=\frac{1}{\lambda_{\max}}$ as $k=\frac{1}{1-n}\log(\frac{\lambda_{\min}}{\lambda_{\max}})$. We arrive from Equation (16) to following equation after substituting all assumptions.

\begin{equation} 
\begin{split}
V\left\{\frac{1}{\lambda}\right\} &= \mathcal{O}\left(\frac{V\{\lambda\}}{M\{\lambda\}}\left(\frac{n-1}{\lambda_{\max}} - \frac{1}{\lambda_{\min}} \right. \right.\\ 
&\quad \left. \left. + \frac{e^{k(1-n)}(kn+1)+k+1}{k^2\lambda_{\min}}\right)\right)
\end{split}
\end{equation}

\begin{equation} 
\begin{split}
V\left\{\frac{1}{\lambda}\right\} &= \mathcal{O}\Bigg(\frac{V\{\lambda\}}{M\{\lambda\}}\Bigg(\frac{n-1}{\lambda_{\max}} - \frac{1}{\lambda_{\min}} \\
&\quad + \frac{1}{\lambda_{\min}}\Bigg(\frac{1}{\Big(\frac{1}{n-1}\log\Big(\frac{\lambda_{\min}}{\lambda_{\max}}\Big)\Big)^2} \\
&\quad \times \Big(\frac{\lambda_{\min}}{\lambda_{\max}}\Big)\Bigg(1+\frac{n}{n-1}\log\Big(\frac{\lambda_{\min}}{\lambda_{\max}}\Big) \\
&\quad +1+\frac{1}{1-n}\log\Big(\frac{\lambda_{\min}}{\lambda_{\max}}\Big)\Bigg)\Bigg)\Bigg)\Bigg)
\end{split}
\end{equation}

The left hand side of equation (18) can be quickly approximated as $\mathcal{O}\left(\frac{V\{\lambda\}}{M\{\lambda\}}\left(\frac{(n-1)\lambda_{\min}-\lambda_{\max}}{\lambda_{\max}\lambda_{\min}}\right)\right)$
. From this, we deduce that the variance of eignenvalues of $\nabla^2\mathcal{L}(\mathcal{R},\theta^*)^{-1}$ is proportional to variance of eigenvalue distribution of $\nabla^2\mathcal{L}(\mathcal{R},\theta^*)$, scaled by reciprocal of its mean. If mean is greater than 1, than the distribution contracts, while the contrary is true otherwise. We take important observation from \cite{ghorbani2019investigation} that outlier eigenvalues of Hessian are usually large, therefore we expect $M\{\lambda\}>1$, even if eigenvalues are more concentrated around zeros. Thus we assert that the variance of eigenvalue distribution of $\nabla^2\mathcal{L}(\mathcal{R},\theta^*)^{-1}$ is less than that of $\nabla^2\mathcal{L}(\mathcal{R},\theta^*)$
\\Furthermore, the distribution becomes more uniform as the difference between maximum and minimum eigenvalue of $\nabla^2\mathcal{L}(\mathcal{R},\theta^*)^{-1}$ is $\frac{\lambda_{\max}-\lambda_{\min}}{\lambda_{\max}\lambda_{\min}}$, so essentially the previous uneven distribution dampens by $\frac{1}{\lambda_{\max}\lambda_{\min}}$.
\\From above conclusion we deduce that $\nabla^2\mathcal{L}(\mathcal{R},\theta^*)^{-1}$ almost acts like a positive scaler, i.e. it almost preserves the direction of the vector it acts upon.

\subsubsection{Orthogonality Condition}
Based on discussion in section 2.2.1, since $\nabla^2\mathcal{L}(\mathcal{R},\theta^*)^{-1}$ approximately preserves the direction of $\nabla\mathcal{L}(\mathcal{F},\theta^*)$, and hence for left hand side of (13) to approach zero, then necessarily and approximately $\nabla\mathcal{L}(\mathcal{R},\theta^*)\cdot \nabla\mathcal{L}(\mathcal{F},\theta^*)\rightarrow{0}$. Thus we find following unlearning orthogonality condition.

\begin{equation} 
\boxed{
\nabla\mathcal{L}(\mathcal{R},\theta^*) \perp \nabla\mathcal{L}(\mathcal{F},\theta^*)
}
\end{equation}

We would want to drop the Hessian since its computation is significantly dominated by gradient computation \cite{guo2019certified}. If in Equation (10), if we instead starting by expanding the loss of unlearning model around pretrained parameters, then the same sequence of steps leads to another orthogonality condition.

\begin{equation} 
\boxed{
\nabla\mathcal{L}(\mathcal{R},\theta) \perp \nabla\mathcal{L}(\mathcal{F},\theta^*)
}
\end{equation}

We derive the \textbf{unlearning metric} that conforms to condition defined in (20) as.

\begin{equation} 
1-\frac{\nabla_{\theta}{\mathcal{L}(\mathcal{R})}\cdot\nabla_{\theta}{\mathcal{L}(\mathcal{F})}}{\|\nabla_{\theta}{\mathcal{L}(\mathcal{R})}\|_2\|\nabla_{\theta}{\mathcal{L}(\mathcal{F})}\|_2}
\end{equation}

%%%%%%%%%%%%%%%%%%%%%%%%%%%%%%%%%%%%%%%%%%%%%%%%%%%%%%%%%%%%%%%%%%%%%%%%%%%%%%%%%%%%%%%%%%%%%%%%%
\begin{figure}[t]
\centering
\hspace*{-0.1in}
\includegraphics[width=6.0cm, height=3cm]{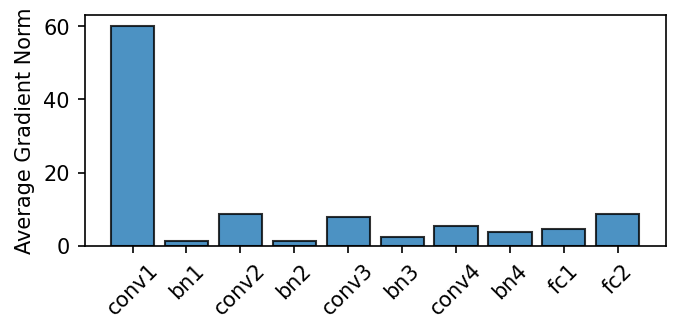}
\caption{Gradient of loss of CNN trained on CIFAR-10 over layers from shallow (left) to deep (right)}
\end{figure}
%%%%%%%%%%%%%%%%%%%%%%%%%%%%%%%%%%%%%%%%%%%%%%%%%%%%%%%%%%%%%%%%%%%%%%%%%%%%%%%%%%%%%%%%%%%%%%%

\section{Gradient of Loss of Neural Network is Dominated by Gradient Over Shallow Layers}

Suppose a model $\mathcal{M}_{\theta}(x)$ with parameters $\theta$ acting on an input $x$ (with corresponding label $y$), can be decomposed into composition of functions as $\mathcal{M}_{\theta}(x)=f_{\theta_1}(g_{\theta_
2}(x))$, where $f$ would represent the deeper layers of model with parameters $\theta_1$, $g$ would represent shallow layers with parameters $\theta_2$, and $\delta\theta_1+\delta\theta_2=\delta\theta$. Essentially, $\theta_1$ is not changed at indices under perturbations, where $g$'s parameters exist and vice versa. Consider following three cases. For sake of simplicity, we represent $\mathcal{L}((\mathcal{M}(x),y),\theta)$ as $\mathcal{L}(\theta)$.
\subsection{Case-1: Perturbations in Shallow Layer's Parameters}
Suppose if we make perturbation in parameters $\theta_2$, possibly in an attempt to train, then the gradient of the loss of the neural network can be written as.
\begin{equation} \label{eq1}
\nabla_\theta{\mathcal{L}(\theta_1,\theta_2+\delta\theta_2)}=\frac{\partial \mathcal{L}}{\partial \mathcal{M}}\cdot\nabla_{\theta}{f_{\theta_1}(g_{\theta_2+\delta\theta_2}(x))}
\end{equation}

We make a Taylor approximation around $\theta_2$, and omitting $o(\|\delta\theta_2\|_2)$ terms,
\begin{equation} \label{eq1}
\nabla_\theta{\mathcal{L}(\theta_1,\theta_2+\delta\theta_2)}=\frac{\partial \mathcal{L}}{\partial \mathcal{M}}\cdot\{\nabla_{\theta}{f_{\theta_1}(g_{\theta_2}(x)+\nabla_{\theta}g(x)\delta\theta_2}\}
\end{equation}

In Equation (24), the $\nabla_{\theta}g(x)$ represents the Jacobian of $g(x,\theta)$ with respect to $\theta$, but for sake of convenience we use same notation as gradient. We make another Taylor approximation of Equation (24) around $g_{\theta_2}(x)$ as.

\begin{equation}
\begin{split}
\nabla_{\theta} \mathcal{L}(\theta_1,\theta_2+\delta\theta_2) &= \frac{\partial \mathcal{L}}{\partial \mathcal{M}} \cdot \left\{ \nabla_{\theta}\left[f_{\theta_1}( g_{\theta_2}(x)) \right. \right.\\
&\quad + \left. \left. \nabla_{\theta}(f(g(x))) \cdot \nabla_{\theta}g(x) \cdot \delta\theta_2(x) \right. \right. \\ 
&\quad + \left. \left. o(\|\nabla_{\theta}g(x) \cdot \delta\theta_2\|_2) \right] \right\}
\end{split}
\end{equation}

\begin{equation} \label{eq1}
\begin{split}
\nabla_{\theta} \mathcal{L}(\theta_1,\theta_2+\delta\theta_2) &= \frac{\partial \mathcal{L}}{\partial \mathcal{M}} \cdot \left\{ \nabla_{\theta}(f_{\theta_1}(g_{\theta_2}(x)) \right.\\
&\quad + \left. \nabla_{\theta}^2(f(g(x))) \cdot \nabla_{\theta}g(x) \cdot \delta\theta_2(x) \right. \\ 
&\quad + \left. \nabla_{\theta}(f(g(x))) \cdot \nabla_{\theta}^2g(x) \cdot \delta\theta_2(x) \right. \\ 
&\quad + \left. o(\|\nabla_{\theta}^2g(x) \cdot \delta\theta_2)\|_2) \right\}
\end{split}
\end{equation}

\subsection{Case-2: Perturbations in Deeper Layer's Parameters}
Likewise to case-1, suppose if we make perturbation in only parameters of $f$ part of the model as.

\begin{equation} 
\nabla_\theta{\mathcal{L}(\theta_1+\delta\theta_1,\theta_2)}=\frac{\partial \mathcal{L}}{\partial \mathcal{M}}\cdot\nabla_{\theta}{f_{\theta_1+\delta\theta_1}(g_{\theta_2}(x))}
\end{equation}

Performing Taylor's expansion around $\theta_1$ in Equation (27).

\begin{equation} 
\begin{split}
\nabla_{\theta} \mathcal{L}(\theta_1+\delta\theta_1,\theta_2) &= \frac{\partial \mathcal{L}}{\partial \mathcal{M}} \cdot \left\{ \nabla_{\theta} f_{\theta_1}(g_{\theta_2}(x)) \right. \\ 
&\quad \left. + \nabla^2_{\theta} f_{\theta_1}(g_{\theta_2}(x)) \cdot \delta\theta_1 + o(\|\delta\theta_1\|_2) \right\}
\end{split}
\end{equation}

%%%%%%%%%%%%%%%%%%%%%%%%%%%%%%%%%%%%%%%%%%%%%%%%%%%%%%%%%%%%%%%%%%%%%%%%%%%%%%%%%%%%%%%%%%%%%%%%%
\begin{figure}[t]
\centering
\hspace*{-0.1in}
\includegraphics[width=8.2cm, height=4.9cm]{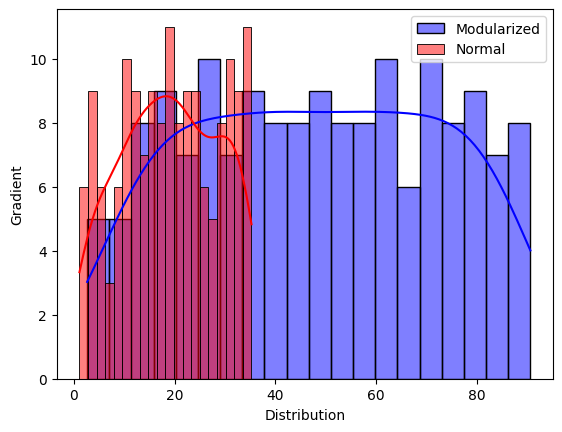}
\caption{Gradient distribution of normed gradient of \textbf{intermediate} in case of Modularized CNN (whose beginning and final are retrained in unlearning phase) and pretrained CNN over retain dataset}
\end{figure}
%%%%%%%%%%%%%%%%%%%%%%%%%%%%%%%%%%%%%%%%%%%%%%%%%%%%%%%%%%%%%%%%%%%%%%%%%%%%%%%%%%%%%%%%%%%%%%%

\subsection{Case-3: Perturbations in Whole Model's Parameters}
Combining case-1 and case-2, where we combine perturbations in $\theta_1$ and $\theta_2$, as.

\begin{equation} \label{eq1}
\nabla_\theta{\mathcal{L}(\theta_1+\delta\theta_1,\theta_2+\delta\theta_2)}=\frac{\partial \mathcal{L}}{\partial \mathcal{M}}\cdot\nabla_{\theta}{f_{\theta_1+\delta\theta_1}(g_{\theta_2+\delta\theta_2}(x))}
\end{equation}

\begin{equation} \label{eq1}
\begin{split}
\nabla_{\theta} \mathcal{L}(\theta_1+\delta\theta_1,\theta_2+\delta\theta_2) &= \frac{\partial \mathcal{L}}{\partial \mathcal{M}} \cdot \left[ \nabla_{\theta} f_{\theta_1}(g_{\theta_2}(x)) \right.\\
&\quad +  \nabla_{\theta}^2 f(g(x)) \cdot \nabla_{\theta} g(x) \cdot \delta\theta_2 \\ 
&\quad +  \nabla_{\theta} f(g(x)) \cdot \nabla_{\theta}^2 g(x) \cdot \delta\theta_2  \\ 
&\quad +  \nabla^2_{\theta} f_{\theta_1}(g_{\theta_2}(x)) \cdot \delta\theta_1 \\
&\quad + o(\|\nabla_{\theta}^2 g(x) \cdot \delta\theta_2\|_2+\|\delta\theta_1\|_2) 
\end{split}
\end{equation}

\begin{table*}[]
\centering
\scriptsize
\setlength{\tabcolsep}{2pt} % default value: 6pt
\renewcommand{\arraystretch}{0.8} % default value: 1
\begin{tabular}{|c|c|c|c|c|c|c|c|}
\hline
         & \textbf{R}        & \textbf{CF}                                                       & \textbf{D}                                                                & \textbf{BD}                                                               & \textbf{S}                                                                 & \textbf{P+U}                                                                & \textbf{MU}                                                                                             \\ \hline
MLP      & \begin{tabular}[c]{@{}l@{}}$\alpha=10^{-3}$\\ bs=256\end{tabular} & \begin{tabular}[c]{@{}l@{}}$\alpha=10^{-3}$\\ bs=256\end{tabular} & \begin{tabular}[c]{@{}l@{}}$\alpha=10^{-3}$\\ bs=256\\ T=4.0 \\hw=1.0 \\sw=$10^{-1}$\end{tabular} & \begin{tabular}[c]{@{}l@{}}$\alpha=10^{-3}$\\ bs=256\\ T=4.0 \\$\text{ratio}_{R}=0.3$\end{tabular} & \begin{tabular}[c]{@{}l@{}}$\alpha=10^{-3}$\\ bs=256\\ $\gamma=10^{-4}$\\$\text{epoch}_{L_1}=15$\end{tabular} & \begin{tabular}[c]{@{}l@{}}$\alpha=10^{-3}$\\ bs=256\\ pr=0.95\end{tabular} & \begin{tabular}[c]{@{}l@{}}$\alpha=10^{-3}$\\ ${\alpha}_3=10^{-4}$\\ $S=30$\\$S_1=15$\\ $\tau=15$\\bs=256\end{tabular} \\ \hline

CNN      & \begin{tabular}[c]{@{}l@{}}lr=$10^{-3}$\\ bs=256\end{tabular} & \begin{tabular}[c]{@{}l@{}}$\alpha=10^{-3}$\\ bs=256\end{tabular} & \begin{tabular}[c]{@{}l@{}}$\alpha=10^{-3}$\\ bs=256\\ T=4.0\\hw=1.0 \\sw=$10^{-1}$\end{tabular} & \begin{tabular}[c]{@{}l@{}}$\alpha=10^{-3}$\\ bs=256\\ T=4.0\\$\text{ratio}_{R}=0.3$\end{tabular} & \begin{tabular}[c]{@{}l@{}}$\alpha=10^{-3}$\\ bs=256\\ $\gamma=10^{-4}$\\$\text{epoch}_{L_1}=15$\end{tabular} & \begin{tabular}[c]{@{}l@{}}$\alpha=10^{-3}$\\ bs=256\\ pr=0.95\end{tabular} & \begin{tabular}[c]{@{}l@{}}$\alpha=3\times10^{-3}$\\ ${\alpha}_3=3\times10^{-4}$\\ $S=30$\\$S_1=10$\\ $\tau=25$\\bs=256\end{tabular} \\ \hline

VGG16    & \begin{tabular}[c]{@{}l@{}}$\alpha=10^{-3}$\\ bs=256\end{tabular} & \begin{tabular}[c]{@{}l@{}}$\alpha=10^{-3}$\\ bs=256\end{tabular} & \begin{tabular}[c]{@{}l@{}}$\alpha=10^{-3}$\\ bs=256\\ T=4.0 \\hw=1.0 \\sw=$10^{-1}$\end{tabular} & \begin{tabular}[c]{@{}l@{}}$\alpha=10^{-3}$\\ bs=256\\ T=4.0\\$\text{ratio}_{R}=0.3$\end{tabular} & \begin{tabular}[c]{@{}l@{}}$\alpha=10^{-3}$\\ bs=256\\ $\gamma=10^{-4}$\\$\text{epoch}_{L_1}=15$ \end{tabular} & \begin{tabular}[c]{@{}l@{}}$\alpha=10^{-3}$\\ bs=256\\ pr=0.95\end{tabular} & \begin{tabular}[c]{@{}l@{}}$\alpha=10^{-3}$\\ ${\alpha}_3=10^{-4}$\\ $S=30$\\$S_1=10$\\ $\tau=15$\\bs=256\end{tabular} \\ \hline

ResNet18 & \begin{tabular}[c]{@{}l@{}}$\alpha=10^{-3}$\\ bs=256\end{tabular} & \begin{tabular}[c]{@{}l@{}}$\alpha=10^{-3}$\\ bs=256\end{tabular} & \begin{tabular}[c]{@{}l@{}}$\alpha=10^{-3}$\\ bs=256\\ T=4.0 \\hw=1.0 \\sw=$10^{-1}$\end{tabular} & \begin{tabular}[c]{@{}l@{}}$\alpha=10^{-3}$\\ bs=256\\ T=4.0\\$\text{ratio}_{R}=0.3$\end{tabular} & \begin{tabular}[c]{@{}l@{}}$\alpha=10^{-3}$\\ bs=256\\ $\gamma=10^{-4}$\\$\text{epoch}_{L_1}=15$\end{tabular} & \begin{tabular}[c]{@{}l@{}}$\alpha=10^{-3}$\\ bs=256\\ pr=0.95\end{tabular} & \begin{tabular}[c]{@{}l@{}}$\alpha=10^{-3}$\\ ${\alpha}_3=10^{-4}$\\ $S=30$\\$S_1=15$\\ $\tau=15$\\bs=256\end{tabular} \\ \hline
\end{tabular}
    \caption{Hyperparameters of benchmark unlearning methods over architectures }
    \label{tab:your_label_here}
\end{table*}

%%%%%%%%%%%%%%%%%%%%%%%%%%%%%%%%%%%%%%%%%%%%%%%%%%%%%%%%%%%%%%%%%%%%%%%%%%%%%%%%%%%%%%%%%%%%%%%%%
\begin{figure*}[t]
\centering
\hspace*{-0.1in}
\includegraphics[width=15.5cm, height=7.1cm]{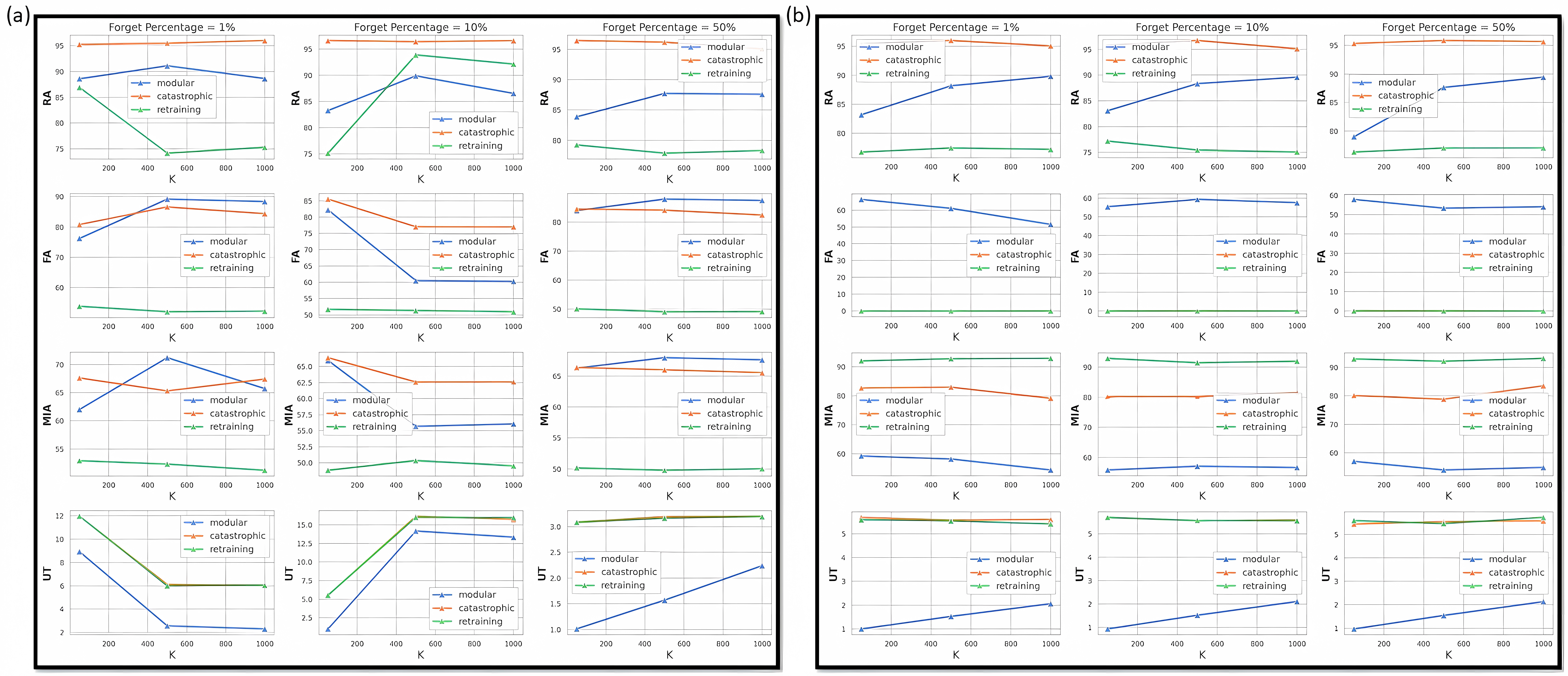}
\caption{(a) Evolution of metrics RA, FA, MIA and UT of modular unlearning over different sizes of clusters K, for the case of arbitrary forgetting, and condensation via fast distribution matching (b) Evolution of metrics RA, FA, MIA and UT of modular unlearning over different sizes of clusters K, for the case of class-wise forgetting, and condensation via fast distribution matching}
\end{figure*}
%%%%%%%%%%%%%%%%%%%%%%%%%%%%%%%%%%%%%%%%%%%%%%%%%%%%%%%%%%%%%%%%%%%%%%%%%%%%%%%%%%%%%%%%%%%%%%%

%%%%%%%%%%%%%%%%%%%%%%%%%%%%%%%%%%%%%%%%%%%%%%%%%%%%%%%%%%%%%%%%%%%%%%%%%%%%%%%%%%%%%%%%%%%%%%%%%
\begin{figure*}[t]
\centering
\hspace*{-0.1in}
\includegraphics[width=15.5cm, height=7.1cm]{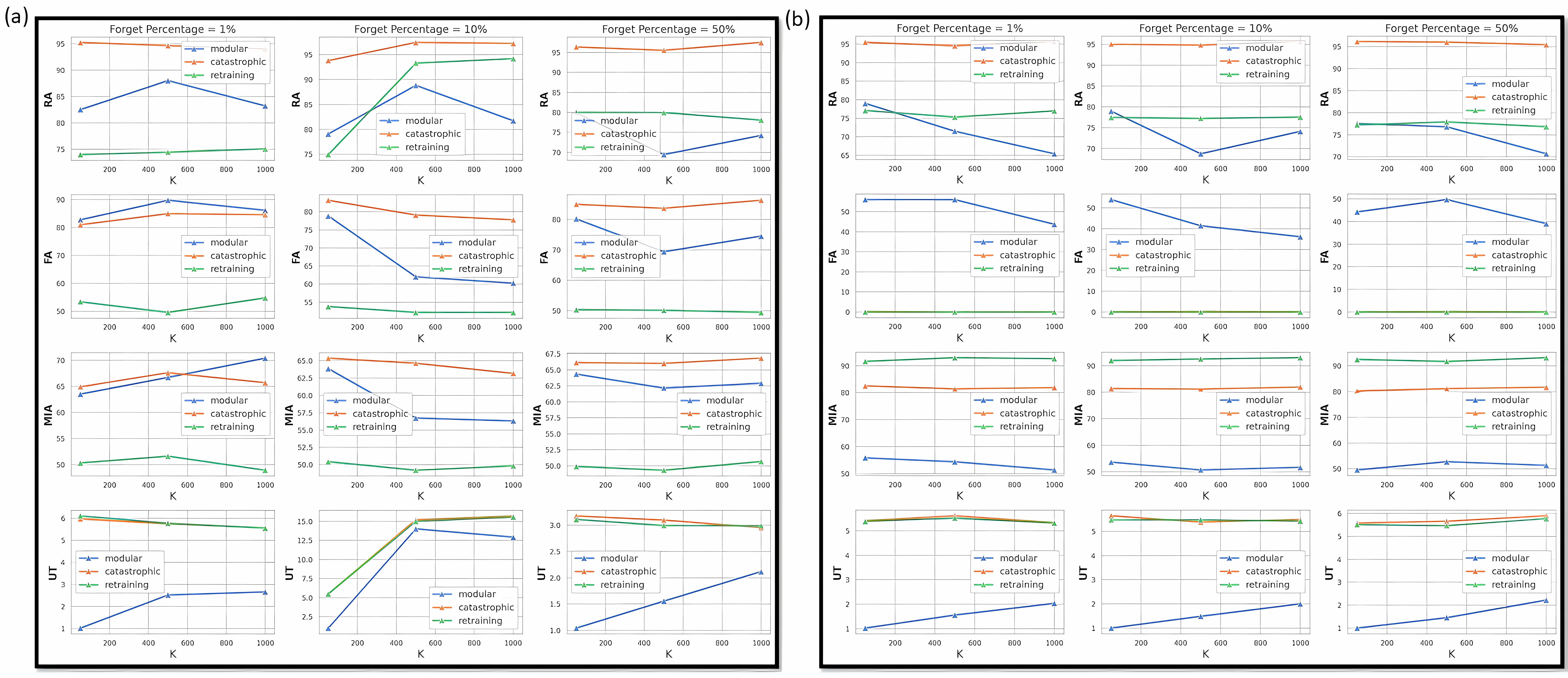}
\caption{(a) Evolution of metrics RA, FA, MIA and UT of modular unlearning over different sizes of clusters K, for the case of arbitrary forgetting, and condensation via model inversion (b) Evolution of metrics RA, FA, MIA and UT of modular unlearning over different sizes of clusters K, for the case of class-wise forgetting, and condensation via model inversion}
\end{figure*}
%%%%%%%%%%%%%%%%%%%%%%%%%%%%%%%%%%%%%%%%%%%%%%%%%%%%%%%%%%%%%%%%%%%%%%%%%%%%%%%%%%%%%%%%%%%%%%%

With the abstractions $f$ and $g$, we can make $g$ sequentially cover from shallow to deep layers to perform an inference.  We first define that $g_i$ covers from the shallowest ($0^{th}$) layer to the $i^{th}$ layer of the model.
Next we see that if we define domain $D$, which contains $\theta_2$ and $\delta\theta_2$, and $\mathbf{M}_i=\sup_{\theta_2\in D}\|\nabla{g_i(x,\theta_2)}\|_2$, then due to \cite{richardson2011advanced}, we can write.
\begin{equation} 
\begin{split}
\|g_i(x,\theta_2+\delta\theta_2)-g_i(x,\theta_2)\|_2\leq{\mathbf{M}_i\|\delta\theta_2\|_2}
\end{split}
\end{equation}
. Since neural networks are randomly initialized with small weights \cite{hanin2018start}, there $\|\delta\theta_2\|_2$ is an invariant quantity, and thus we write Equation (31) as.
\begin{equation} 
\begin{split}
\|g_i(x,\theta_2+\delta\theta_2)-g_i(x,\theta_2)\|_2=\mathcal{O}(\mathbf{M}_i)
\end{split}
\end{equation}
Now we use a key observation from \cite{petzka2021relative} that represents that multiplicative perturbations in parameters as equivalent to multiplication perturbations in inputs (features). Hence we use this fact to rewrite Equation (26) as.
\begin{equation} 
\begin{split}
\|g_i(x+\delta{x},\theta_2)-g_i(x,\theta_2)\|_2=\mathcal{O}(\mathbf{M}_i)
\end{split}
\end{equation}

Next we take a general observation about neural networks that shallower layers of neural network learn low level features of the input, and thus sensitive to input \cite{mo2020darknetz, dosovitskiy2016inverting}. Especially in convolution neural networks, the layer representations achieve more translation invariance as we move deeper \cite{wiatowski2017mathematical}. Hence, with this observation we establish an equivalence using Equation (33) that as $i$ moves from $0$ to $l$ layers, the left hand side progressively decreases, and simultaneously $\mathbf{M}_i$ progressively becomes smaller. We illustrate the effect in figure 3, where the norm of gradient of layers of trained CNN are shown, averaged over CIFAR-10 training dataset.

\par
Thus we finally make inference from Equation (30) that $\|\nabla_{\theta} \mathcal{L}(\mathcal{M}(x), y, \theta)\|_2$ is dominated by terms containing $\|\nabla_{\theta} g(x)\|_2$ and $\|\nabla_{\theta}^2 g(x)\|_2$ as $\|\delta_1\|_2, \|\delta_2\|_2 \rightarrow{0}$, since $\delta\theta_1$ and $\delta\theta_2$ are arbitrary and independent.
Based on our previous inference on trend of Jacobian over layers of model, we deduce that gradient based perturbations (which can be in the form of unlearning) over the whole model's parameters is dominated with the perturbations in shallower layers, while deeper layer parameters are less impacted. This effects serves as one of the main reasons for the principle of dividing the network into abstractions of \textbf{beginning}, \textbf{intermediate} and \textbf{final}, and training \textbf{beginning} during proposed unlearning phase. By changing the parameters of \textbf{beginning} towards new minima over retain dataset, leads to rapid forgetting in \textbf{intermediate}, which can be achieved with even 1 epoch. This stratagey is depicted in figure 2, where retraining the whole network leads to smaller gradients in \textbf{intermediate} part, as compared to our strategy (modularized).

\section{Asymptotic Expression of $\eta_R$}
Assuming that $|\Gamma_{ij}|=|\Gamma_{ik}|$ for $j\neq{k}$, i.e. each cluster is of same size, and the forget samples are uniformly distributed through out the cluster. Through applications of collection protocol, total count of reduced retain dataset is.

\begin{equation} \label{eq1}
\begin{split}
N_r= T_1+T_2
\end{split}
\end{equation}
where $T_1$ is the count of cluster, not containing any forget samples, while $T_2$ is the count of the retain samples from clusters, that do contain the forget samples.
Assuming that the forget samples are uniform randomly distributed through out the dataset, then the expected count of retain images found in forget-infected clusters is given by.

\begin{equation} \label{eq1}
\begin{split}
T_2=\sum_{\phi_{ij}}\sum_{m=0}^{|\phi_{ij}|-1}\binom{|\phi_{ij}|-1}{m}(\frac{1}{cK})^m(1-\frac{1}{cK})^{|\phi_{ij}|-1-m}\\(\frac{N}{cK}-1-m)
\end{split}
\end{equation}

Here the inner sum represents the expect number of retain samples achieved from clusters, affected by the forget samples, while the outer sum accumulated the expected number per cluster over all clusters. $\phi_{ij}\subseteq\Gamma_{ij}$ is the affected portion of the cluster by forget samples, and hence $\sum_{\phi_{ij}}=\sum_{i,j}|\phi_{ij}|=N_F$ and $|\phi_{ij}|\leq\frac{N}{cK}$. We can create an asymptotic bound for Equation (35) by substituting $|\phi_{ij}|$ as $\frac{N}{cK}$, and then summing up the outer sum, we get.

\begin{equation} \label{eq1}
\begin{split}
T_2=\mathcal{O}(\sum_{m=0}^{\frac{N}{S}-1}\binom{\frac{N}{cK}-1}{m}(\frac{1}{cK})^m(1-\frac{1}{cK})^{\frac{N}{cK}-1-j}\\(\frac{N}{cK}-1-m)\sum_{\phi_{ij}})
\end{split}
\end{equation}

\begin{equation} \label{eq1}
\begin{split}
T_2=\mathcal{O}(\sum_{m=0}^{\frac{N}{cK}-1}\binom{\frac{N}{cK}-1}{m}(\frac{1}{cK})^m(1-\frac{1}{cK})^{\frac{N}{cK}-1-j}\\(\frac{N}{cK}-1-m)|N_F|)
\end{split}
\end{equation}

On the other hand, we calculate the expectation of $T_1$ as follows.
\begin{equation} \label{eq1}
\begin{split}
T_1=p_{miss}K
\end{split}
\end{equation}
We can calculate $p_{\text{miss}}$ as follows. If the probability of 1 forget sample is in arbitrary one the $cK$ clusters is $\frac{|\Gamma_{ij}|}{N}$, then probability that the 1 forget sample is not in one of the $cK$ clusters is $1-\frac{\Gamma_{ij}}{N}$. We can extend this probability of $N_F$ forget samples not in one of the $cK$ clusters as $(1-\frac{\Gamma_{ij}}{N})^{N_F}$. Since $\Gamma_{ij}=\frac{N}{cK}$, as the number of samples in each cluster is same by assumption, therefore $(1-\frac{1}{cK})^{N_F}$, which is in fact $p_{\text{miss}}$. Henceforth, 
\begin{equation} \label{eq1}
\begin{split}
T_1=(1-\frac{1}{cK})^{N_F}K
\end{split}
\end{equation}

Combining Equations (36) and (39) into Equation (34), we get.
\begin{equation} 
\begin{split}
N_r=(1-\frac{1}{cK})^{N_F}K+\mathcal{O}(\sum_{m=0}^{\frac{N}{S}-1}\binom{\frac{N}{cK}-1}{m}\\(\frac{1}{cK})^m(1-\frac{1}{cK})^{\frac{N}{cK}-1-j}\\(\frac{N}{cK}-1-m)N_F)
\end{split}
\end{equation}

Expressing $N_F=N_D-N_R$, and representing $\eta_R=\frac{N_r}{N_R}$, we can express Equation (40) as.
\begin{equation}
\begin{split}
\eta_R=(1-\frac{1}{cK})^{N-}\frac{K}{R}+\mathcal{O}(\sum_{m=0}^{\frac{N}{cK}-1}\binom{\frac{N}{cK}-1}{m}\\(\frac{1}{cK})^m(1-\frac{1}{cK})^{\frac{N}{cK}-1-j}\\(\frac{N}{cK}-1-m)(N_D-N_R))
\end{split}
\end{equation}

%%%%%%%%%%%%%%%%%%%%%%%%%%%%%%%%%%%%%%%%%%%%%%%%%%%%%%%%%%%%%%%%%%%%%%%%%%%%%%%%%%%%%%%%%%%%%%%%%
\begin{figure}[t]
\centering
\hspace*{-0.1in}
\includegraphics[width=8.0cm, height=7.4cm]{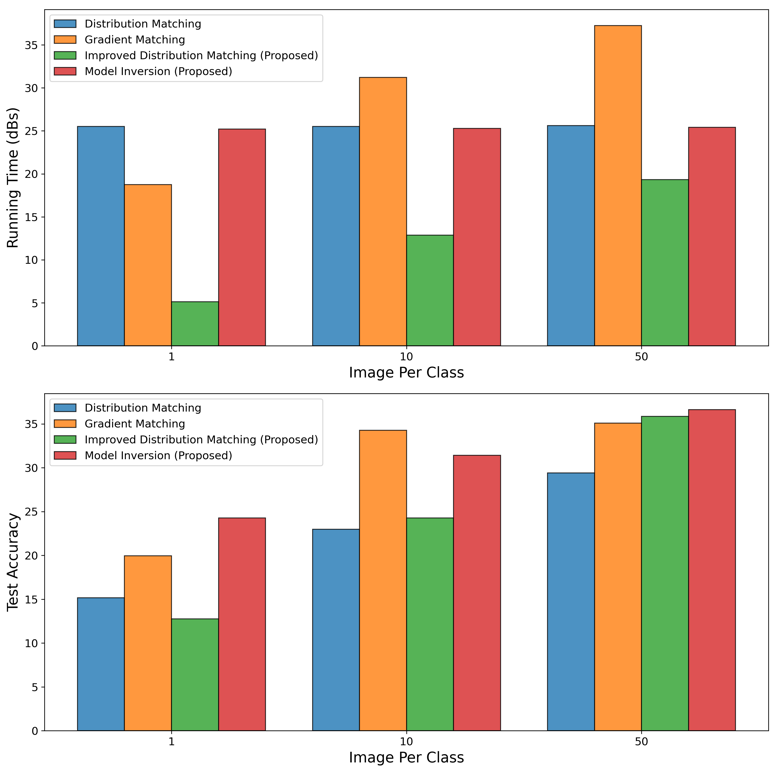}
\caption{Benchmarking of dataset condensation of CIFAR10 dataset via Gradient Matching (GM), Distribution Matching (DM), and our proposed Improved Distribution Matching (IDM) and Model Inversion (MI) based condensation, while the dataset contains 5000 images per class }
\end{figure}
%%%%%%%%%%%%%%%%%%%%%%%%%%%%%%%%%%%%%%%%%%%%%%%%%%%%%%%%%%%%%%%%%%%%%%%%%%%%%%%%%%%%%%%%%%%%%%%

%%%%%%%%%%%%%%%%%%%%%%%%%%%%%%%%%%%%%%%%%%%%%%%%%%%%%%%%%%%%%%%%%%%%%%%%%%%%%%%%%%%%%%%%%%%%%%%%%
\begin{figure*}[t]
\centering
\hspace*{-0.1in}
\includegraphics[width=15.5cm, height=6.1cm]{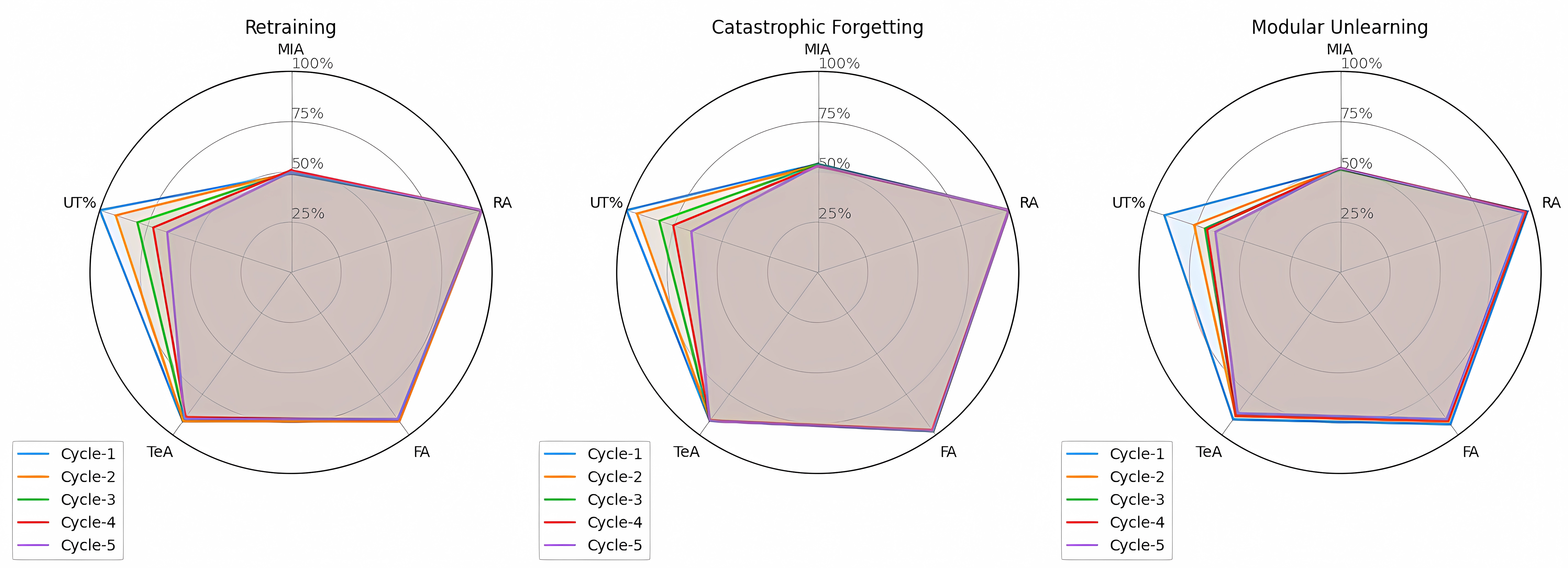}
\caption{Evolution of \textbf{RA}, \textbf{FA}, \textbf{TrA} and \textbf{UT} (represented as percentage here by taking ratio of unlearning time over all three methods) of unlearning via retraining, catastrophic forgetting and modular unlearning over 5 unlearning cycles with forgetting 10 percent of dataset per cycle}
\end{figure*}
%%%%%%%%%%%%%%%%%%%%%%%%%%%%%%%%%%%%%%%%%%%%%%%%%%%%%%%%%%%%%%%%%%%%%%%%%%%%%%%%%%%%%%%%%%%%%%%

Through application of binomial theorem, and simplification, we arrive at following expression.

\begin{equation} 
\boxed{
\begin{aligned}
\eta_R &= (1 - \frac{1}{cK})^{N_D - cK} \frac{cK}{N_R} \\
&\quad + \mathcal{O}\left(\left(\frac{N_D}{N_R} - 1\right)\left(1 - \frac{1}{cK}\right)\left(\frac{N_D}{cK} - 1\right)\right)
\end{aligned}
}
\end{equation}

Under the same assumptions, we realize the scenario where collection protocol leads to $\eta_R=1$, equivalent to coupon collector's problem \cite{blom1993problems}. Under this problem, the expected number of forget samples to hit all the clusters is.
\begin{equation}
    N_F=cK\sum_{i=1}^{cK}\frac{1}{i}
\end{equation}

We use a well known sharp inequality $\sum_{i=1}^{cK}\frac{1}{i}\geq\log(cK)+\frac{1}{cK}$ to achieve a prophylactic equality in equation (43). Thus we arrive at.
\begin{equation}
    N_F=cK\log(cK)+1
\end{equation}

By substituting $N_F=N_D-N_R$ and replacing the equality with an inequality for $\eta_R$ to be less than 1 under expectation, then necessarily.

\begin{equation}
\boxed{
    N_R > N_D- cK\log(cK) + 1
    }
\end{equation}

\section{Intuition Behind Overfitting Metric}
We take a key observation from \cite{carlini2022membership}, where the distribution of losses of model over training dataset are more closer to zero mean, while the distribution of losses over test dataset are further away from zero mean. Through a strong connection between overfitting samples and membership inference attack, overfitted model (which achieve small over loss over training dataset) are more susceptible to inferring the presence or absence of random sample belonging to training dataset. more For a model trained on dataset $\mathcal{D}$, it achieves a local minima of its associated loss with stationary condition $\nabla_{\theta}\mathcal{L}(\mathcal{D},\theta)=0$. Thus we capture the model achieving minima over dataset, as well as overfitting on it when $\nabla_{\theta}\mathcal{L}(\mathcal{D},\theta)$ and $\mathcal{L}(\mathcal{D},\theta)$ both approach zero. Consequently, we capture this essence through following overfitting metric with values in $\mathbb{R}$, such that smaller values would imply more overfitting per sample input-output pair $(T_i,l_i)\in\mathcal{D}$.
\begin{equation}
|(\mathcal{L}((T_i,l_i),\theta))-\text{mean}(|\nabla_{\theta}\mathcal{L}((T_i,l_i),\theta)|)|
\end{equation}

\section{Hyperparameters for Unlearning Benchmarking}

Throughout our experiments, we utilized our proposed fast distribution matching based dataset condensation for progression of our modularized unlearning, if not specified. For the offline phase of modular unlearning framework, we set the $L=10$ iterations, $L_1=20$ and $L_2=20$ iterations with corresponding learning rate $\alpha_1=10^{-4}$ and $\alpha_2=10^{-5}$ respectively.
\par
The MLP comprises of 3 linear layers with ReLU activation. The CNN comprises of 4 convolution layers with batch normalization and ReLU activation, then max-pooling operation, finally a 2-layered MLP with dropout operation. The VGG16 and ResNet18 architecture is according to \cite{simonyan2014very} and \cite{he2016deep} respectively.
\\The remaining hyperparameters used in our experimentation are summarized in table-1, where the hyperparameters of modular unlearning are described in online phase. For all the learning algorithms, the training iterations `$\text{epochs}_{\text{main}}$' are set to 30 for experimentation over random sample forgetting, and 10 for case of single class-forgetting. $\alpha$ generally means the learning rate associated with training associated with unlearning, $T$ stands for the temperature associated with distillation based training, $\text{bs}$ stands for size of batch during training. Other hyperparameters like $\text{ratio}_R$ stands for the ratio of randomly sample retain dataset utilized, $\text{pr}$ stands for pruning ratio associated with model prunning algorithm, and $\text{epoch}_{L_1}$ stands for the limit of of total unlearning epochs from end, till which $\gamma$ associated with parameter $L_1$ regularization is linearly decayed from its initial value via relation $\gamma(1-\frac{\text{current epoch}}{\text{epochs}_{\text{main}}-\text{epoch}_{L_1}})$, otherwise $\gamma$ is thresholded to zero \cite{jia2023model}.

% \newpage
\section{Experiments of Evolution of Unlearning Performance over Variations in $K$}
We consider two separate experimentations, where we we use both of our proposed dataset condensation schemes, i.e Fast Distribution Matching and Model Inversion based, for the offline phase of retain dataset reduction framework. With choice of dataset as CIFAR-10, we consider consider $K=50$, $K=500$ and $K=1000$, while the CIFAR-10's training dataset has 5000 images per class. Furthermore, we also impose condition that if $\eta_R>0.7$, then $\text{epochs}_{\text{main}}=30$. On the contrary, if $0.4<\eta_R\leq 0.7$, then $\text{epochs}_{\text{main}}=20$, and lastly if $\eta_R\leq 0.4$, then $\text{epochs}_{\text{main}}=10$. This allows to preserve unlearning utility for our unlearning algorithm, when higher compression leads to reduction in accuracy during prolonged training.
\\The results are shown in figure 3 and figure 4, where fast distribution matching and model inversion based dataset condensation was utilized respectively. \textbf{MU}'s unlearning time increases as $K$ increases (because $\eta_R$ increases due to equation (42), although not directly applicable as k-means leads to clusters of unequal sizes). Nevertheless, the performance of unlearning remains stable with variations in $K$, showing the potential of proposed unlearning to minimize retain dataset, and still have good unlearning performance.

%%%%%%%%%%%%%%%%%%%%%%%%%%%%%%%%%%%%%%%%%%%%%%%%%%%%%%%%%%%%%%%%%%%%%%%%%%%%%%%%%%%%%%%%%%%%%%%%%
\begin{figure*}[t]
\centering
\hspace*{-0.1in}
\includegraphics[width=15.5cm, height=6.1cm]{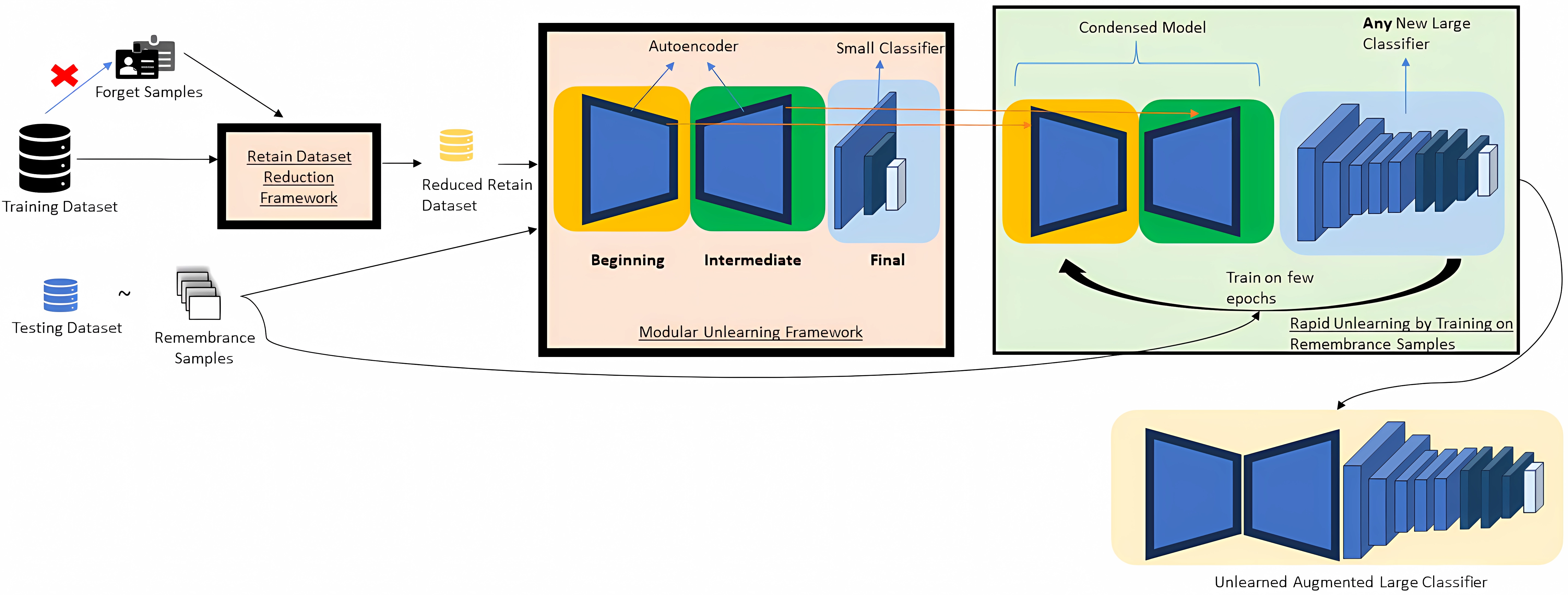}
\caption{Proposed methodology of utilization of proposed unlearning methodology in the scenario of unlearning in condensation}
\end{figure*}
%%%%%%%%%%%%%%%%%%%%%%%%%%%%%%%%%%%%%%%%%%%%%%%%%%%%%%%%%%%%%%%%%%%%%%%%%%%%%%%%%%%%%%%%%%%%%%%

\section{Benchmarking of Proposed Dataset Condensation Techniques}
We benchmark our dataset condensation with two main dataset condensation techniques, namely Gradient Matching \cite{zhao2020dataset} and Distribution matching \cite{zhao2023dataset}. The results are shown in figure 5, where we attempt to condense 5000 images per class of CIFAR-10 into 1, 10 and 50 synthetic images. It is vivid from top and bottom bar graph that proposed approaches are either equivalent or better in performance, but with advantage of faster condensation time, especially for our fast distribution matching based approach.

\section{Experiment over Multiple Unlearning Cycles}
We explore the case where unlearning happens over several unlearning cycles; for example, every month a new batch of data is deleted. We show the unlearning performance of the proposed unlearning with retraining, and catastrophic forgetting, over 5 unlearning cycles with \(10\%\) training deletion per cycle, in Figure 6. While modular unlearning maintains metrics like \textbf{RA}, \textbf{FA}, and \textbf{TrA} (Training accuracy), it shows utility capability in parallel to retraining and catastrophic forgetting, with the advantage of lesser unlearning time for at least the first 4 cycles. However, larger unlearning cycles lead to a smaller retained dataset size for the same \(K\), and thus through Equation (42), \(\eta_R\) increases, leading to the same unlearning time as that of retraining or catastrophic forgetting.

\section{More Detailed Explanation on Unlearning in Condensation as Application of Proposed Unlearning Scheme}
Unlearning in dataset condensation is non-trivial, because if the whole training dataset is condensed into few images per class, then removing the information of few samples of training dataset can involve using the whole retain dataset in optimizing the condensed images, at least with techniques like gradient matching \cite{zhao2020dataset} and distribution matching \cite{zhao2023dataset}. Not to mention that it is not guaranteed to unlearn, unlike the similar notion of catastrophic forgetting which can have some mathematical base for small distance between learned and unlearned parameters, because dataset condensation is so far based on heuristic ideas. To this end, we apply our proposed unlearning methodology in dataset condensation, by view the condensed dataset as a `condensed model' with remembrance samples, which represent as unlearned condensed dataset. We create this correspondence by observing that this condensed model based proxy satisfies two importance properties of condensed dataset. First, it can be quickly used to train any new random model with considerably less amount of time, as compared to training on the original dataset. Secondly, the gained accuracy from this procedure is equivalent to that of the original dataset.
\par
We exploit this correspondence into unlearning of forget samples from the condensed model (equivalent to removing of forget samples from original dataset), which can any new large image classification architecture. The strategy to achieve this illustrated in figure 7. Reduced retain dataset is constructed by using information of forget samples to be unlearning. The \textbf{beginning} and \textbf{intermediate} are assigned an auto-encoder architecture, and \textbf{final} is assigned some small classifier model. By application of modular unlearning over reduced retain dataset on this setup, we replace \textbf{final} with the target large architecture, and then train this modified setup with remembrance samples, leading to unlearned target architecture, augmented with the autoencoder.
\par
The significance of this gets highlighted when we note that size of remembrance samples is very small, for example for CIFAR-10 case, we chose 10 images per class as remembrance samples, while original dataset comprised of 5000 images per class. Therefore, we achieve very fast training of large architectures, much significantly faster than the original training or even achievable through condensed dataset, with almost equivalent accuracy. This is characterized with learning without knowledge of forget samples, but at the cost of increase of parameter count in the target architecture due to addition of autoencoder. These results can witness from our experiments.

\end{document}